\documentclass[manuscript,screen]{acmart}
\usepackage{xspace}
\usepackage{graphicx}
\usepackage{booktabs}
\usepackage{multirow}
\usepackage{caption}
\usepackage{array}

\usepackage{tcolorbox}
\usepackage{adjustbox}
\usepackage{subcaption}
\usepackage{colortbl}
\AtBeginDocument{%
  }

\setcopyright{acmlicensed}
\copyrightyear{2025}
\acmYear{2025}
\acmDOI{XXXXXXX.XXXXXXX}

\acmISBN{978-1-4503-XXXX-X/18/06}

\newcommand{\projectName}{NeMo\xspace}

\begin{document}

\title{\projectName: A Neuron-Level Modularizing-While-Training Approach for Decomposing DNN Models }

\author{Xiaohan Bi}
\affiliation{
  \institution{State Key Laboratory of Complex \& Critical Software Environment, Beihang University}
  \city{Beijing}
  \country{China}
}
\affiliation{
  \institution{Peng Cheng Laboratory}
  \city{Shenzhen}
  \country{China}
}
\email{xhbii@buaa.edu.cn}

\author{Binhang Qi}
\affiliation{
\institution{National University of Singapore}
\country{Singapore}}
\affiliation{
\institution{Beihang University}
\city{Beijing}
\country{China}
}

\email{qibh@nus.edu.sg}

\author{Hailong Sun}
\authornote{Corresponding author}
\affiliation{
 \institution{State Key Laboratory of Complex \& Critical Software Environment, Beihang University}
 \city{Beijing}
 \country{China}
}
\affiliation{
 \institution{Hangzhou Innovation Institute of Beihang University}
 \city{Hangzhou}
 \country{China}
}

\email{sunhl@buaa.edu.cn}

\author{Xiang Gao}
\affiliation{
 \institution{State Key Laboratory of Complex \& Critical Software Environment, Beihang University}
 \city{Beijing}
 \country{China}
}
\affiliation{
 \institution{Hangzhou Innovation Institute of Beihang University}
 \city{Hangzhou}
 \country{China}
}
\email{xiang_gao@buaa.edu.cn}

\author{Yue Yu}
\affiliation{
  \institution{Peng Cheng Laboratory}
  \city{Shenzhen}
  \country{China}
}
\email{yuy@pcl.ac.cn}

\author{Xiaojun Liang}
\affiliation{
  \institution{Peng Cheng Laboratory}
  \city{Shenzhen}
  \country{China}
}
\email{liangxj@pcl.ac.cn}

\renewcommand{\shortauthors}{Bi et al.}

\begin{abstract}

With the growing incorporation of deep neural network (DNN) models into modern software systems, the prohibitive construction costs of DNN models have become a significant challenge in software development. To address this challenge, model reuse has been widely applied to reduce model training costs; however, indiscriminately reusing an entire model may incur significant inference overhead. Consequently, DNN modularization --- borrowing the idea of modularization in software engineering --- has increasingly gained attention, enabling module reuse by decomposing a DNN model into modules. In particular, the emerging modularizing-while-training (MwT) paradigm, which outperforms modularizing-after-training by incorporating modularization into the model's training process, has been demonstrated as a more effective approach for DNN modularization. However, existing MwT approaches focus on small-scale convolutional neural network (CNN) models at the convolutional kernel level. They struggle to handle diverse DNNs and large-scale models, particularly Transformer-based models, which consistently achieve state-of-the-art results across various tasks.

To address these limitations, we propose \projectName, a scalable and more generalizable MwT approach. \projectName operates at the neuron level --- a fundamental component common to all DNNs --- thereby ensuring applicability to Transformers and various DNN architectures. Moreover, we design a contrastive learning-based modular training method, equipped with an effective composite loss function, hence being scalable to large-scale models. Comprehensive experiments on two Transformer-based models and four CNN models across two widely-used classification datasets demonstrate \projectName's superiority over the state-of-the-art MwT method. Results show average performance gains of 1.72\% in module classification accuracy and a 58.10\% reduction in module size. Our findings demonstrate that \projectName exhibits efficacy across both CNN and large-scale Transformer-based models.
Moreover, a case study based on open-source projects demonstrates the potential benefits of \projectName in practical scenarios, offering a promising approach for achieving scalable and generalizable DNN modularization.

\end{abstract}




\maketitle

  \newcommand{\nbc}[3]{
        {\colorbox{#3}{\scriptsize\textcolor{white}{#1}}}
            {\textcolor{#3}{\small$\blacktriangleright$\textit{#2}$\blacktriangleleft$}}
}

\newcommand{\change}[1]{{\color{red}#1}
}
\newcommand{\xiang}[1]{\nbc{XG}{#1}{blue}}

\section{Introduction}

The increasing integration of deep neural network (DNN) models into contemporary software systems~\cite{pei2017deepxplore,qi2021dreamloc} has rendered DNN training a crucial component of the software development lifecycle. However, the training process for DNNs, particularly those with billions of parameters and large datasets, can be prohibitively expensive. 
To mitigate development and training costs, model reuse~\cite{engineering2030,taraghi2024deep,davis2023reusing,transfer_survey,transferable} has been widely adopted in practice, drawing significant attention from both the AI and software engineering communities.
Researchers have explored techniques such as transfer learning~\cite{bert, decaf, transfer_survey, transferable} and model retrieval~\cite{automrm} to facilitate effective model reuse while developing engineering best practices---such as model reengineering~\cite{jiang2024challenges} and optimizing the model supply chain~\cite{wang2024large}---to enhance efficiency and safety.
However, these approaches typically treat DNN models as monolithic artifacts, and indiscriminate reuse of entire models can incur additional inference overhead and introduce security vulnerabilities.

DNN models and software programs share fundamental similarities ~\cite{DNNPan,pei2017deepxplore}, with the former encoding functionalities through data-driven, automatic training processes, and the latter implementing functionalities via explicit coding.
In software engineering, program development typically adheres to the principle of modularity, facilitating on-demand reuse of program components ~\cite{ModularSE, ModularSE1, ModularSE2}. 
Borrowing the idea of modularization in software engineering, DNN model modularization and on-demand model reuse are receiving increasing attention.
For example, Pan et al. and Qi et al. pioneered DNN modularization on fully connected neural networks (FCNNs)~\cite{DNNPan, SeaM}, convolutional neural networks (CNNs)~\cite{CNNPan, CNNSplitter, GradSplitter, SeaM, MwT, modelfoundry} and recurrent neural networks (RNNs)~\cite{iowaRNN} models by identifying the relevant weights or special structural components for each functionality of the model. 
Each module possesses a part of the model's functionalities, retaining only the weights responsible for its specific function.
For instance, a 10-class image classification model can be decomposed into 10 modules, each retaining only the weights responsible for recognizing one single class.
Beyond model reuse, DNN modularization offers potential benefits for model maintenance~\cite{DeepArc}, deployment~\cite{davis2023reusing}, and supply chain management~\cite{wang2024large} by identifying the relevant modules.

According to when the modularization is performed, existing DNN modularization approaches can be categorized into \textit{modularizing-after-training}~\cite{CNNSplitter, GradSplitter, SeaM, NUSxinchao2022deep, iowaRNN, zhang2022remos, DeepArc} and \textit{modularizing-while-training}~\cite{MwT}, with the former decomposing a trained DNN model and the latter integrating modularization into the training progress of a randomly initialized model and then decomposing the modular trained model.
In our previous work~\cite{MwT}, we proposed a modular training method, MwT, which integrates the two factors of cohesion and coupling into the training loss.
``Cohesion'' evaluates the overlap between the sets of convolutional kernels used for the same class of samples, and ``coupling'' assesses the overlap between the sets of convolutional kernels used for different classes of samples.
By optimizing these factors, MwT minimizes the overlap between weight sets responsible for different functionalities. 
Since the modular model is specifically trained for modularization, \textit{modularizing-while-training} significantly outperforms \textit{modularizing-after-training} in module size, inference performance, and the modularization time cost.
However, incorporating modularization into the training process is very challenging, as it needs to consider the training details of various architectures and scales of models and avoid side effects on both model performance and training time cost. 
Our experimental observations reveal that even the state-of-the-art \textit{modularizing-while-training} approach, MwT~\cite{MwT}, is far from practical application due to generalizability issues in \textit{model architecture} and \textit{model size}.

Current DNN modularization techniques, including MwT, are primarily constrained to small-scale models, such as FCNN and CNN models with up to only 14.7 million parameters (trainable parameters used for modularizing the model). 
As Transformer-based models continue achieving state-of-the-art results on image ~\cite{ViT, deit, SwinTransformer}, natural language ~\cite{transformer, GPT4}, and code-related tasks ~\cite{codebert, codellama}, they have largely supplanted traditional CNN and RNN architectures in numerous applications.
The superiority of Transformer-based models is not only derived from advanced architecture ~\cite{transformer} but also benefits from large-scale parameters ~\cite{GPT4}. 
On the other hand, such large-scale parameters incur a surge in training and inference costs, thus a DNN modularization approach for large-scale Transformer-based models is more necessary.
Although MwT presents a general framework applicable to diverse DNN models, its concrete implementation is limited to CNN modularization and does not support Transformer-based models. 
The main reasons may include: (1) MwT performs modularization at the convolutional kernel level, a structural component specific to CNNs, and (2) MwT faces performance scalability challenges with large-scale models. It utilizes a straightforward summation of multiple losses (i.e., cohesion and coupling losses) for evaluation and optimization. Such a manner has been proven, through theoretical analysis~\cite{multitask1, multitask2, multitask3} and experiments (see Table~\ref{tab:rq1_vit_mwt_ours}), to be inefficient for optimization.

Addressing the aforementioned limitations presents significant challenges. First, \textit{Choosing appropriate modularization granularity is difficult.} Although MwT could be adapted to Transformers by changing the convolutional kernel level to the special structural component of Transformers, such as attention head level, the fact that the number of attention heads is typically small makes it infeasible to divide attention heads into numerous modules and achieve both high module performance and low overlap between modules.
For example, a Vision Transformer (ViT) ~\cite{ViT} model fine-tuned on the CIFAR10 dataset has only 96 attention heads. Assuming that we decompose it into ten modules with each corresponding to one class and enforce no overlap between modules, each module would contain around 10 attention heads. 
Given there are 12 attention layers in the ViT model, it means that the two layers did not contain any attention heads, which is unacceptable.
In contrast, even small CNN models possess a substantial number of convolutional kernels, such as the VGG16 for 10-class classification has 4226 kernels, making modularization on convolutional kernel level feasible. 
Although the existing work~\cite{SeaM, CNNPan, DNNPan} that performs modularization at the weight level provides an alternative idea, such modularization granularity can aggravate the second challenge.
Second, \textit{Designing an effective loss function incorporating multiple objectives is difficult}. The training loss should involve three optimization objectives, including inference performance (e.g., cross-entropy loss), cohesion, and coupling. Summation of the three parts directly is a straightforward way; however, such a manner has been proven to be ineffective for optimization through theoretical analysis~\cite{multitask1, multitask2, multitask3}. 
Our empirical findings indicate that this challenge is further amplified as the scale of model weights and modularization-related trainable parameters increases. The effective modularization on a large scale of models necessitates a more sophisticated training loss function.

To deal with the challenges above, we propose \projectName, a pioneering NEuron-level MOdularizing-while-training approach.
Unlike existing methods that implement modularization at the weight ~\cite{SeaM} or special structural component level ~\cite{CNNSplitter, GradSplitter, MwT}, \projectName operates at the neuron level. This level of granularity effectively addresses the first challenge and offers three advantages over MwT: 
(1) Neurons are the foundational components of DNN models, enabling \projectName to be seamlessly applied to Transformer architectures and easily extended to other DNNs.
(2) The number of neurons in a model lies between the number of weights and special structural components, facilitating a balance between modularity and inference performance. 
(3) Similar to convolutional kernels, irrelevant neurons can be physically removed from modules. Also, as a neuron contains fewer weights than a convolutional kernel, \projectName is a finer granularity approach than MwT, thus performing better in modularization.
To address the second challenge, we optimize the cohesion and coupling loss functions by introducing the technique of contrastive learning ~\cite{MoCo, SimCLR, InfoDistribute}. 
The improved loss function incorporates inference performance, cohesion, and coupling in exponential and fractional forms, instead of simply adding them together, thus achieving better modularization performance and module inference performance.
Additionally, the improved loss function has only a single hyperparameter while three hyperparameters for MwT, making \projectName easier to adapt to different models.

We conducted comprehensive experiments to evaluate \projectName, utilizing two Transformer-based models (ViT and DeiT) and four mainstream CNN architectures across two widely-used image classification datasets. To ensure a fair comparison, we also adapted MwT to operate at the neuron level for Transformer-based models.
Experimental results demonstrate \projectName's efficacy in enabling modular training and decomposition in both Transformer-based and CNN models. Compared to MwT, \projectName not only improves the modular training in accuracy but also significantly reduces the Neuron Retention Rate (NRR) or Kernel Retention Rate (KRR). For the ViT model, \projectName achieves up to a 64.76\% reduction in NRR compared to MwT. Regarding on-demand model reuse, \projectName achieves a significant improvement in reducing model size, with an average weight reduction of 57.85\% compared to MwT.
Moreover, a case study based on the open-source repositories~\cite{spoof_detect,bone_detection,rock_paper_scissors} further demonstrates the practical benefits of \projectName in real-world scenarios. 
In this case study,
\projectName allows developers to reuse only the relevant module from a pre-trained model, thereby reducing the inference overhead of the fine-tuned model on their downstream tasks.

The main contributions of this work are as follows:
\begin{itemize}
    \item To the best of our knowledge, \projectName is the first neuron-level modularizing-while-training approach to support modular training and structured decomposition for Transformer-based models.
    \item We propose a contrastive learning-based method for optimizing cohesion and coupling, which makes modular training of large-scale models feasible.
    \item We have conducted extensive experiments using two representative Transformer-based and four CNN models on two widely-used datasets. The results demonstrate that \projectName can outperform the state-of-the-art approaches in module classification accuracy and module size. 
    Moreover, we have conducted a case study to demonstrate the practical benefits of \projectName in real-world scenarios.
    We have published the replication package of \projectName \cite{bi2024nemo}.
\end{itemize}

\section{Background}
This section briefly introduces some preliminary information about this study, including contrastive learning (CL) and some mainstream neural network models. 
\subsection{Mainstream neural network models}
\label{subsec:bg_vit}
Neural networks~\cite{AlexNet, DenseNet, RNN} are computational models consisting of interconnected nodes (neurons) organized in layers, designed to learn complex patterns from data. Their fundamental structure consists of an input layer, one or more hidden layers, and an output layer. During the learning process, the weights associated with inter-neuron connections are adjusted. 
\textbf{Convolutional Neural Networks (CNNs)}~\cite{CNN1, CNN2, CNN3} represent a specialized class of neural networks optimized for processing grid-like data, particularly images. CNNs introduce convolutional layers that apply filters to input data, enabling the network to capture local patterns and spatial hierarchies. Key components of CNNs include: 
\textit{convolutional layers}, which extract features using learnable filters;
\textit{pooling layers}, which reduce spatial dimensions and computational complexity; and
\textit{fully connected layers}, which perform high-level reasoning based on extracted features.

The \textbf{Vision Transformer (ViT)}~\cite{ViT} model adapts the transformer architecture, originally designed for natural language processing, to computer vision tasks. ViT divides an image into fixed-size patches, linearly embeds these patches, and processes them with a standard transformer encoder. Key components of ViT include:
\textit{patch embedding}, which converts image patches into linear embeddings;
\textit{positional encoding}, which adds information about the spatial position of patches;
\textit{multi-head self-attention}, which allows the model to attend to different parts of the input; and
\textit{feedforward networks}, which process the attention output.

ViT has demonstrated exceptional performance in image classification tasks, often surpassing CNN-based models with minimal domain-specific adaptations. Building upon this success, Touvron et al. (2021) proposed Data-efficient image Transformers (DeiT) ~\cite{deit} to address the data efficiency limitations of the original ViT model.

Existing structured modularization approaches~\cite{CNNSplitter, GradSplitter, MwT} primarily target convolutional kernels within CNN models. These methods typically estimate the significance of each kernel by analyzing its output channels, subsequently removing irrelevant kernels during the decomposition stage. However, this methodology is not applicable to Transformer-based models, which fundamentally differ in their structure and propagation (see Section ~\ref{sec:modularization}).

\subsection{Contrastive Learning}
Contrastive Learning and its recent developments aim to train encoders that capture shared information representations across different parts of high-dimensional signals ~\cite{InstDis, InfoNCE}. The core idea of contrastive learning is to pull together similar samples (positive pairs) while pushing apart dissimilar samples (negative pairs). This method is especially valuable in self-supervised learning settings as it does not rely on manually labeled data.

To enable the encoder to fully learn the features, Wu et al.~\cite{InstDis} introduced a large memory bank to store the feature representations of all samples in the dataset. He et al.~\cite{MoCo} introduced Momentum Contrast (MoCo), which addresses the dynamic nature of negative samples by maintaining a queue of negative examples and using a momentum encoder to ensure consistency in feature representations. Chen et al.~\cite{SimCLR} simplify the contrastive learning framework by removing the need for memory banks or specialized architectures. Instead, it uses large batch sizes and data augmentation strategies to generate positive and negative pairs on-the-fly. Caron et al.~\cite{SwAV} proposed SwAV, a method that combines contrastive learning with clustering. Grill et al.~\cite{BYOL} introduced BYOL, a method that eliminates the need for negative samples altogether. 
The cohesion and coupling metrics employed in MwT ~\cite{MwT} quantify the similarity of selected neurons within the same category and across different categories, respectively. It shares fundamental similarities with various contrastive learning techniques but lacks efficient optimization loss for models with numerous neurons.
Thus, we incorporate contrastive learning into modular training and take labels into account to improve the calculation of cohesion and coupling.

\section{Approach}
This section details the methodology of \projectName. \projectName aims at modular training and decomposing an \textit{n-class} classification model into several modules, each containing a subset of neurons and functions. 

\subsection{Overview of \projectName}
\begin{figure}[t]
\centerline{\includegraphics[width=0.9\linewidth]{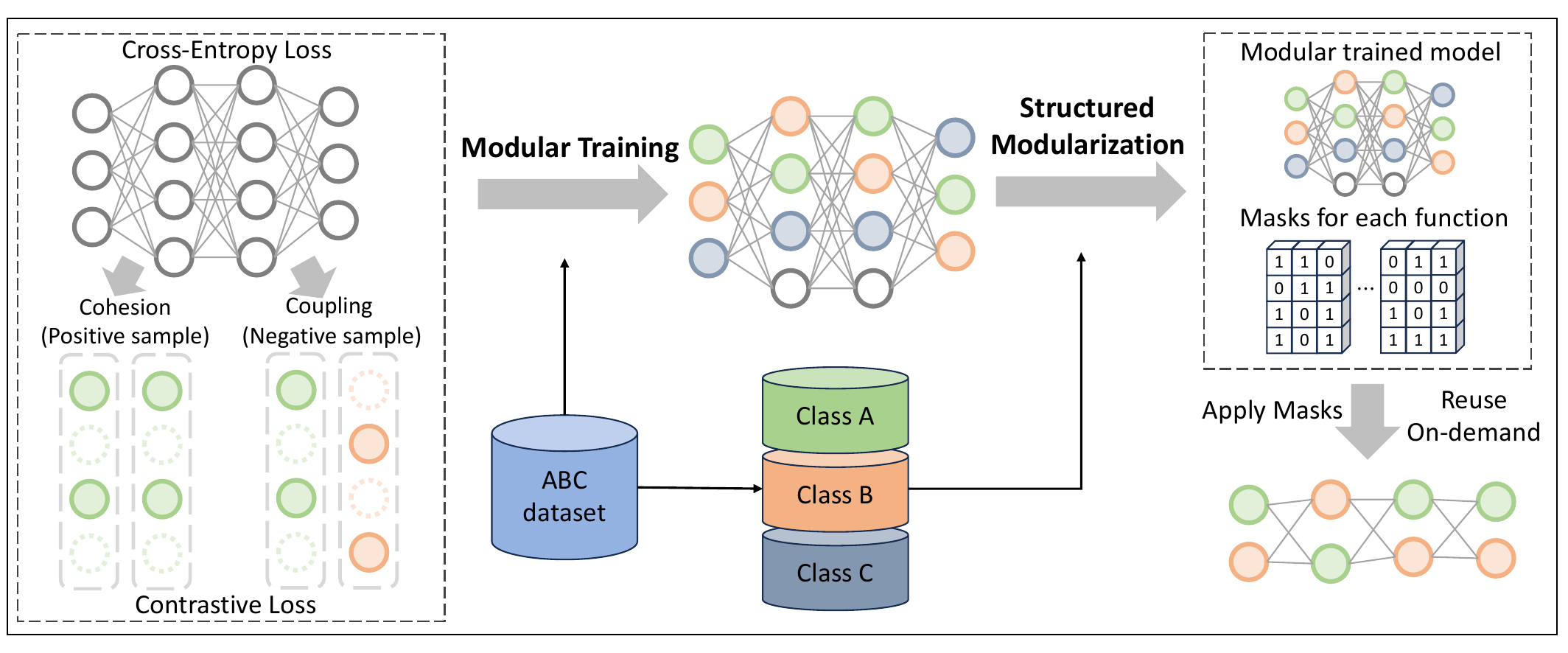}}
\caption{The overall framework of \projectName.}
\label{fig:fig_1}
\vspace{-8pt}
\end{figure}

As shown in Figure~\ref{fig:fig_1}, for a randomly initialized model, \projectName achieves on-demand reuse through two phases.
(1) \textit{Modular training}. \projectName begins with a randomly initialized model and incorporates a \textit{neuron identifier}, which consists of several \textit{mask generators}. To achieve modular training, we introduce a combined loss function that includes both accuracy and modularity loss. During the training process, \projectName increases the cohesion of the model and reduces its coupling by applying cross-entropy and contrastive learning algorithms.  
(2) \textit{Structured Modularization}. After the training process, \projectName generates masks for each functionality and then obtains modules by removing irrelevant neurons structurally from the modular trained model according to the masks.
Figure~\ref{fig:fig_2} provides a detailed workflow of \projectName. Specifically, during the modular training phase, to identify relevant neurons, a neuron identifier is attached to the model. It comprises several mask generators, each corresponding to a network layer. The generated masks represent the relevance of neurons to specific subtasks. These masks are utilized to compute the contrastive loss for model optimization and the subsequent pruning process.
Upon completion of model training, \projectName employs these masks to physically remove redundant neurons from each layer. This process effectively prunes specific weights from the weight matrices, resulting in a tailored submodel. The functionality of this submodel is precisely aligned with user requirements, ensuring that it retains only the essential components necessary for the specified subtask while eliminating extraneous elements. 

\subsubsection{Recognize relevant neurons}

\begin{figure}[t]
\centerline{\includegraphics[width=1\linewidth]{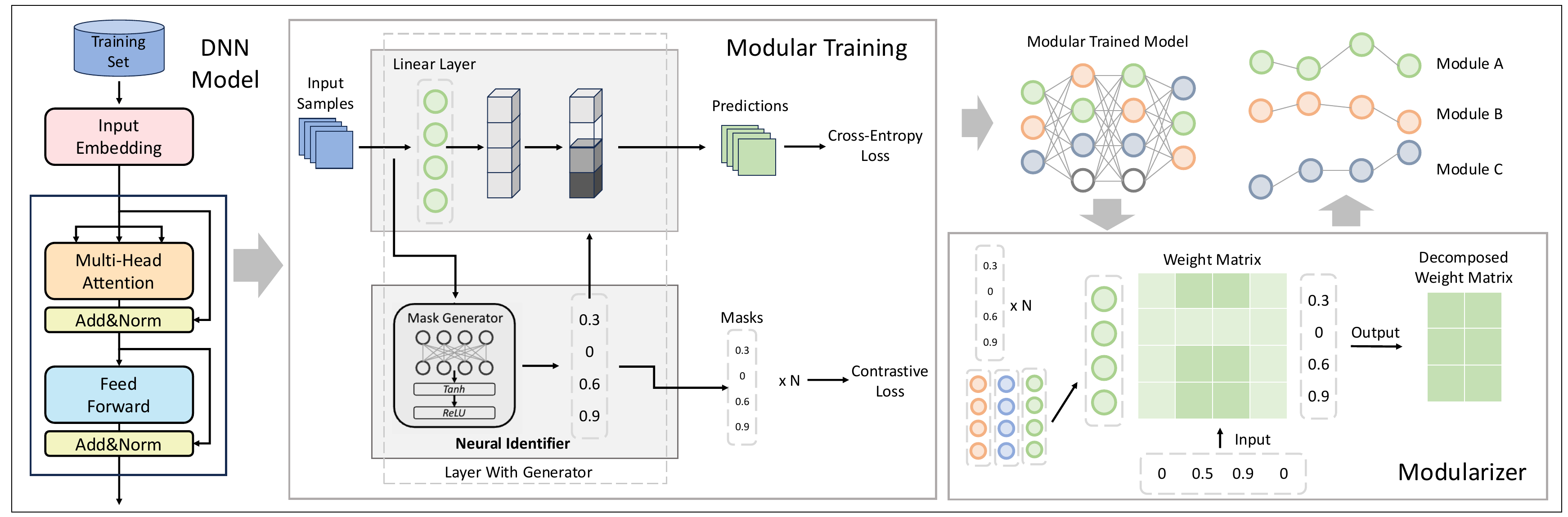}}
\caption{The workflow of \projectName.}
\label{fig:fig_2}
\end{figure}

A crucial process in modular training is identifying which neurons are responsible for each class.
As shown in Figure \ref{fig:fig_2}, \projectName incorporates a \textit{neuron identifier} into the DNN model (e.g., Transformer-based models).
The \textit{neuron identifier} is trained jointly with the DNN model, learning to identify the neurons responsible for each class (see Sections \ref{sec:3.2.3} and \ref{sec:3.2.4}).
The \textit{neuron identifier} consists of \textit{mask generators} (denoted as $G$), each corresponds to a linear layer (denoted as $l_{DNN}$) in the DNN model.
Each $G$ shares the same input as its associated $l_{DNN}$ and learns to identify the neurons responsible for the input. 
The output of $G$ is a mask $m$, a vector where each element corresponds to a neuron in $l_{DNN}$.
Each element of $m$ lies in the range $[0,1)$, representing the likelihood that the corresponding neuron is responsible for the input.
To remove irrelevant neurons during training, \projectName applies $m$ to the output of $l_{DNN}$ via element-wise multiplication, zeroing out the outputs of the neurons deemed irrelevant.

More specifically, the process of generating a mask $m$ and applying it to the output $h$ of $l_{DNN}$ to obtain the masked output $\tilde{h}$ can be formulated as follows:
$$
h = l_{DNN}(x),  \quad h \in \mathbb{R}^{B \times N \times D}
$$
$$
x_{pooled} = AveragePool(x), \quad x_{pooled} \in \mathbb{R}^{B \times 1 \times D}
$$
$$
m = G(x_{pooled}), \quad m \in \mathbb{R}^{B \times 1 \times D}
$$
$$
\tilde{h} = h \odot m, \quad \tilde{h} \in \mathbb{R}^{B \times N \times D}
$$

\noindent To mitigate significant computational overhead, we apply average pooling to the raw input $x$, generating a downsampled input $x_{pooled}$ with reduced dimensions before feeding it into the mask generator $G$.
The \textit{mask generator} $G$ encodes $x_{pooled}$ and outputs a mask $m$, representing which neurons in $l_{DNN}$ are responsible for the input sample. 
As illustrated in Figure ~\ref{fig:fig_2}, each layer in the modular training process comprises the $l_{DNN}$ and $G$.
Each $l_{DNN}$ and its \textit{mask generator} $G$ receive input tensors $x$ with dimensions $(B, N, D)$, where $B$ represents the batch size of input data. In the context of the ViT model
, $N$ represents the number of patches, and $D$ is the feature size per patch (see Section ~\ref{subsec:bg_vit}). For instance, an input image of size 224×224 is typically divided into 14×14 patches (i.e., $N=196+1$, another patch is for global feature), with each patch having a size of 16×16 (i.e., $D=256$)~\cite{ViT}. 
During the forward propagation, the training data is fed into the DNN model and \textit{neuron identifier}. While $l_{DNN}$ outputs the feature $h$, the neuron identifier generates a mask $m$ for each $l_{DNN}$ with dimensions $(B, 1, D)$, corresponding to each neuron's output. The mask generator employs $Tanh$ and $ReLU$ activation functions, constraining mask values to the range $[0,1)$. A value of $0$ indicates that the corresponding neuron's output is irrelevant to the input sample. 

\subsubsection{Evaluating the Performance of Modularity and Classification}
\label{sec:3.2.3}
In modular training, Qi et al.~\cite{MwT} introduced the concepts of cohesion and coupling to assess the modularity performance. Cohesion quantifies the overlap in convolution kernel usage within the same class of samples, whereas coupling measures this overlap across different classes. 
To optimize cohesion and coupling, MwT~\cite{MwT} computes the similarity between the masks of samples and integrates the similarity into the loss function by directly adding it with the cross-entropy loss. However, there are some drawbacks to this loss calculation. First, it treats all samples equally, ignoring the varying difficulty levels between different samples. Second, directly calculating and summing the losses can lead to numerical instability and difficulty in optimization~\cite{multitask1, multitask2, multitask3}. Additionally, the distribution of samples is not controlled, making it difficult for the identifier to learn certain features~\cite{InstDis}.
Consequently, the evaluation and optimization for modularity necessitates a more sophisticated algorithm.

Contrastive learning (CL) seeks to develop useful representations by differentiating between similar and dissimilar data point pairs. Conventional CL methods ~\cite{SimCLR, MoCo, BYOL, SwAV, Barlow_Twins}, such as SimCLR~\cite{SimCLR} and MoCo~\cite{MoCo}, typically involve the creation of positive pairs by applying data augmentations to the same instance, while negative pairs are generated from different instances within the batch. These methods leverage a contrastive loss function, such as the InfoNCE loss~\cite{InfoNCE}, to maximize the similarity between positive pairs and minimize the similarity between negative pairs. Then, they have a neural network encoder with well-trained features for image classification or other downstream tasks.

Drawing inspiration from the concepts of cohesion and coupling, as well as contrastive learning, we propose a supervised contrastive learning approach for modular training. We employ a \textit{neuron identifier} to determine neuron selection, aiming to differentiate neurons based on their specific functional responsibilities. 
More formally, assuming a dataset contains $n$ classes, for class $c_i$, the $n_i$ samples belonging to class $c_{i}$ are denoted as $\{s_{i}^{1}, s_{i}^{2},...,  s_{i}^{n_i}\}$, and the corresponding group of neurons responsible for each sample is represented as $\{sN_{i}^{1}, sN_{i}^{2},..., sN_{i}^{n_i}\}$. Consequently, the module $mN_{i}$, responsible for class $c_i$, consists of all the neurons used for the samples, and is calculated by $\cup_{j=1}^{n_i}sN_{i}^{j}$. In MwT, cohesion measures the extent of overlap in neuron usage among similar samples. The cohesion of the module $mN_i$ responsible for the class $c_i$ is calculated as follows:
\begin{gather}
    Cohesion(m_i) = \frac{2}{n_i \times (n_i-1)} \times \sum_{0<j<k\leq n_i} Overlap(sN_i^j, sN_i^k). \label{eq:cohesion}
\end{gather}
The cohesion calculates the overlap among the samples with the same class, which could be considered as the overlap of positive samples. And the coupling between $m_i$ and $m_j$ is calculated as follows:
\begin{gather}
    Coupling(m_i, m_j) = Overlap(mN_i, mN_j). \label{eq:coupling}
\end{gather}
The overlap metric is derived using the Jaccard Index, which is widely used~\cite{MwT, CNNSplitter, CNNPan} to measure the overlap between modules. For two sets A and B, their Jaccard Index is calculated as:
\begin{gather}
    JI(A, B) = \frac{|A \cap B|}{|A \cup B|}. \label{eq:jaccard}
\end{gather}

\subsubsection{Modular Training and Optimization}
\label{sec:3.2.4}
To optimize modularity performance during training, Eqs. \ref{eq:cohesion} and \ref{eq:coupling} need to be transformed into differentiable loss functions. We employ cosine similarity to calculate the overlap between neuron groups. The cohesion loss for a module $m_i$ and the coupling loss for a pair of modules $(m_i, m_j)$ are as follows:
\begin{gather}
    \mathcal{L}_{cohesion}(m_i) = \frac{2}{n_i \times (n_i-1)} \times \sum_{0<j<k\leq n_i} Cos(sM_i^j, sM_i^k), \label{eq:cohesion_cos}
\end{gather}
\begin{gather}
    \mathcal{L}_{coupling}(m_i, m_j) = \frac{1}{n_i \times n_j} \times \sum_{k=1}^{n_i}\sum_{h=1}^{n_j} Cos(sM_i^k, sM_j^h), \label{eq:coupling_cos}
\end{gather}
Here, $n_i$ represents the number of samples in each batch belonging to class $c_i$, $sM_i^j$ denotes the mask of the $j$-th sample of class $c_i$, and $Cos(a, b)$ is the cosine similarity between tensors $a$ and $b$.
For the masks $sM_i^j$ produced by the neuron identifier, the usual gradient descent optimization algorithms are only effective for optimizing continuous values, not discrete values. Therefore, we represent the selection of neurons by a continuous mask. 
With mask element values in the range $[0,1)$, the cosine similarity between two masks falls within $[0,1]$.

Unlike MwT, which computes the overall cohesion loss $\mathcal{L}_{cohesion}$ and coupling loss $\mathcal{L}_{coupling}$ by averaging the results across all (pairs of) modules and then obtains the final loss by adding them directly, we introduce a temperature parameter $\tau$ and an exponential function to scale the sample distribution~\cite{CL1, CL2}. This modification enables the neuron identifier to learn features more effectively \cite{Multiview}. We compute the overall cohesion loss as follows:
\begin{gather}
     \mathcal{L}_{cohesion} = \frac{1}{n} \times \sum_{0 \leq i \leq n} exp \left( \frac{\mathcal{L}_{cohesion}(m_i )}{\tau} \right).  
     \label{eq:pos_score}
\end{gather}
The overall coupling loss is computed as follows:
\begin{gather}
     \mathcal{L}_{coupling} = \frac{2}{n \times (n-1)} \times \sum_{0 \leq i < j \leq n} exp \left( \frac{\mathcal{L}_{coupling}(m_i, m_j)}{\tau} \right).  
     \label{eq:neg_score}
\end{gather}
After calculating the loss of cohesion and coupling improved by contrastive learning according to Eq.~\ref{eq:pos_score} and Eq.~\ref{eq:neg_score}, the contrastive loss is defined as the proportion of the cohesion loss to the total loss:
\begin{gather}
\mathcal{L}_{contra} = -\log \left( \frac{\mathcal{L}_{cohesion}}{\mathcal{L}_{cohesion} + \mathcal{L}_{coupling}} \right).
\end{gather}

Furthermore, to optimize the performance in classification, we apply the cross-entropy~\cite{AlexNet, cross_entropy} loss $\mathcal{L}_{ce}$ to improve the accuracy of classification. With the contrastive loss function and the cross-entropy function, the objective loss function is defined as:
$$
\mathcal{L} = \mathcal{L}_{ce} + \alpha \times \mathcal{L}_{contra},
$$
where $\alpha$ denotes the weighting factor of contrastive loss. 
Based on the designed loss function, the model is trained through gradient descent so that it learns to use the corresponding sets of neurons (i.e., modules) to recognize different classes of samples.

\subsection{Structured Modularization}
\label{sec:modularization}
After modular training, we decompose the modular model using neuron masks generated by the neuron identifier. Specifically, we first determine what structure the neuron corresponds to in the model and then decompose it according to neuron masks.
\subsubsection{Determine the Structure of Neurons}
In the convolutional layer, a neuron typically represents a convolutional kernel and its receptive field on the input feature map, producing an output element in one of the output feature map channels. Given the complexity of the design, we form an ensemble of neurons from a certain channel for modular training, i.e., a convolutional kernel as an independent substructure. For the removal of convolutional kernels, refer to MwT~\cite{MwT}.

In linear (or fully connected) layers, neurons aggregate weighted inputs from all preceding neurons, add a bias term, and apply an activation function. These neurons primarily conduct linear transformations and feature combinations, achieving nonlinear mapping via the bias term ~\cite{neuron1}. Unlike the neurons in convolutional layers, neurons in linear layers do not constitute a special structure (e.g., convolution kernel). Unlike convolutional layer neurons, those in linear layers lack a specific structural unit (e.g., convolution kernel). To effect neuron removal, we propose a novel decomposition method that eliminates portions of the weight matrix in the linear transformation operation.

\subsubsection{Decompose Linear Layer.}

\begin{figure}[t]
\centerline{\includegraphics[width=1\linewidth]{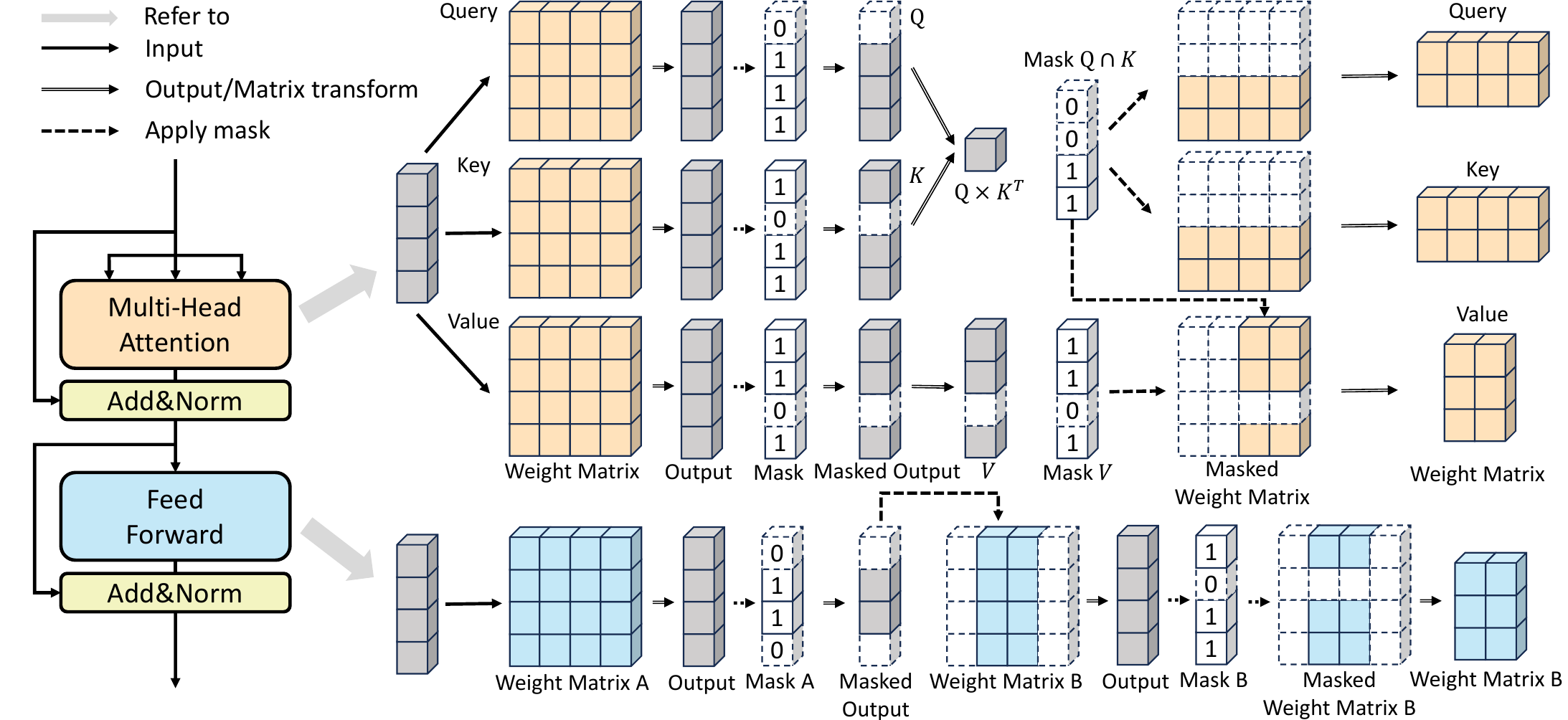}}
\caption{The process of removing neurons in attention and feed-forward layers.}
\label{fig:fig_3}
\end{figure}
To decompose the modular model into modules, \projectName first generates neuron masks for each sample. For class $c_i$, the $n_i$ samples belonging to class $c_{i}$ are denoted as $\{s_{i}^{1}, s_{i}^{2},... s_{i}^{n_i}\}$, and their corresponding masks are denoted as $\{M_{i}^{1}, M_{i}^{2},..., M_{i}^{n_i}\}$. We define:

$$\left. Bin(x) = \left\{\begin{array}{l}0, x\leq 0,\\1,x > 0.\end{array}\right.\right.$$

For class $c_i$, to select useful neurons for all samples and filter the noise, the module mask $M_i = \{\frac{neuron\_count}{n_i}>threshold:1, 0 \mid neuron\_count\in \sum_{j = 1}^{n_i}Bin(M_{i}^{j})\}$. 

The masks determine which neurons in linear layers should be retained or removed.
In a linear layer, each neuron corresponds to an output channel, i.e., a row in the weight matrix. 
To remove irrelevant neurons, we remove the output channels corresponding to the zero-valued elements in the mask.
For example, consider a linear layer implemented as \texttt{nn.linear(20, 5)} with a mask $[1, 0, 1, 0, 0]$.
A masked version of this layer with irrelevant output channels removed would be \texttt{nn.linear(20, 2)}, where its weight matrix retains only the first and third rows of the original weight matrix.

In Transformer models, all layers---including MLP and attention layers---are fundamentally composed of linear operations. 
Therefore, the removal of output channels is applied uniformly across all such layers. 
Additionally, the input channels of each linear layer must be adjusted to match the modified output channels of its previous layer.
For MLP layers, where the output of one layer is directly input to the next, we remove the input channels of the subsequent layer based on the mask of the previous layer.
As for attention layers, updating input channels requires considering the computational dependencies among the query (Q), key (K), and value (V) matrices. Accordingly, we apply the appropriate masks to remove irrelevant input channels based on these relationships, illustrated in ~\autoref{fig:fig_3}. 

\textbf{MLP Layers.}
In each layer's linear mapping, neuron removal is achieved by removing weights corresponding to output dimensions as per the mask. However, this process becomes complex for adjacent layers.  After removing the weight matrix for layer A in Figure~\ref{fig:fig_3}, the output dimension has changed and cannot be aligned with layer B. While zero-padding layer A's output could achieve alignment, it would render part of layer B's input dimension redundant. To optimize computational efficiency, we instead prune layer B's input dimensions based on layer A's mask.

\textbf{Attention Layers.} In the ViT model, the attention layer's structure differs significantly from the feed-forward layer. In Figure ~\ref{fig:fig_3}, the query, key, and value layers similarly process the input $x$ and output $Q$, $K$, and $V$. Then, the attention output is:
$$Attention(Q,K,V)=softmax(\frac{QK^T}{\sqrt{d_k}})V,$$
where $Q = [ q_1, q_2,...,q_n ] $, $K = [ k_1, k_2,...,k_n ]$, and $QK^T = \sum_{i=1}^{n}q_i \times k_i $.
When $q_i$ is masked (i.e., its output becomes zero), the product $q_i \times k_i$ is consequently zero. Therefore, any mask applied to $Q$ should be correspondingly applied to $K$.
Therefore, query and key layers should share a common neuron mask, derived from the intersection of their individual masks. Consider the module mask $m_Q$ for the query layer and $m_K$ for the key layer. Their common mask for output dimension is $m_{QK} = m_Q \cap m_K$. The output dimension of the value layer can be easily removed by $m_V$. Furthermore, the input dimensions of all query, key, and value layers are removed based on the output mask of the previous feed-forward layer output.

\begin{figure}[t]
\centerline{\includegraphics[width=0.7\linewidth]{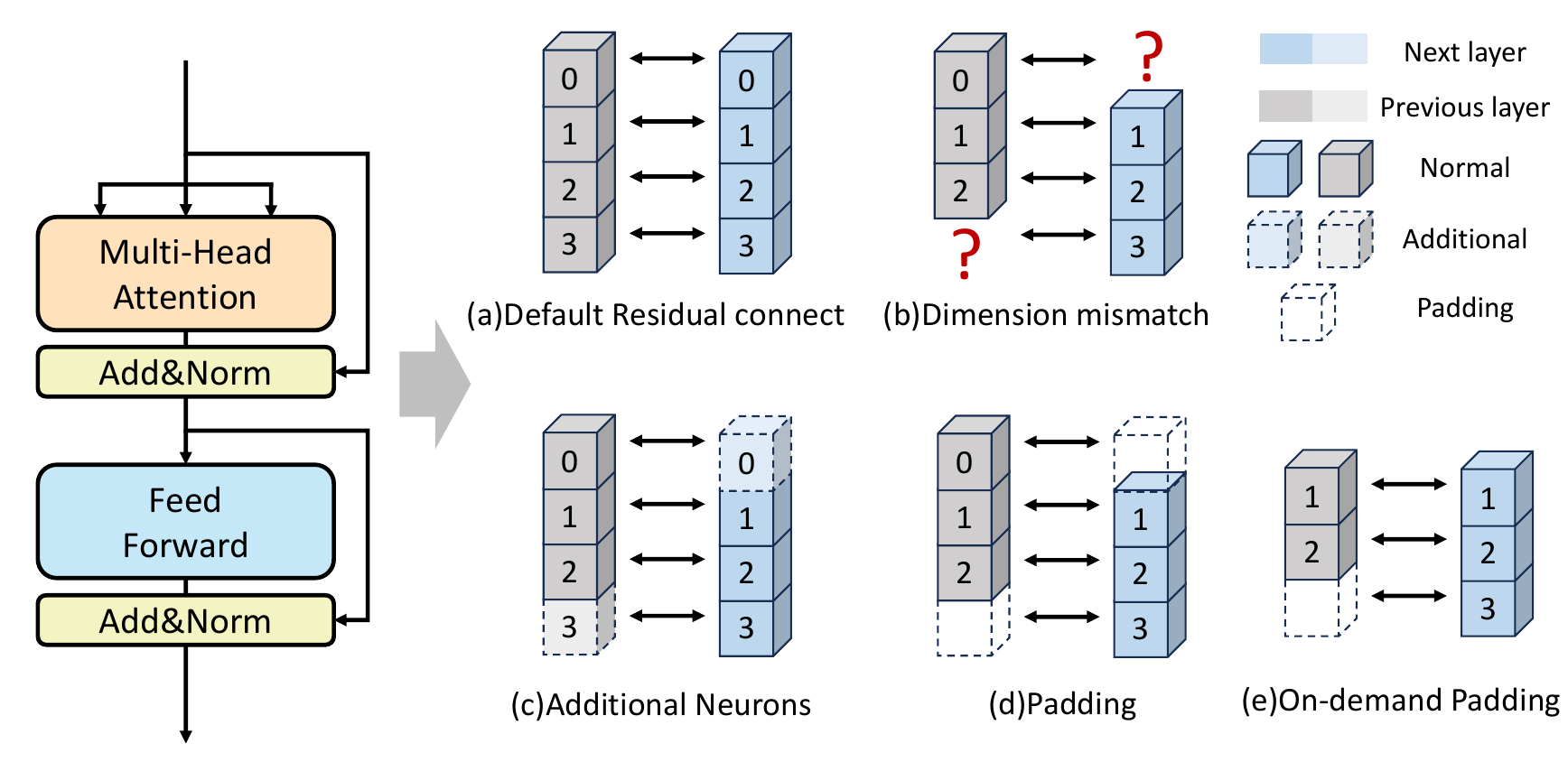}}
\caption{Residual connection mismatch and padding.}
\label{fig:fig_4}
\end{figure}

\textbf{Residual Connection.}
Residual connections ~\cite{residual} between network layers introduce complexities in the modularization process. The selective pruning of the weight matrix can result in dimensional misalignment between the input and output of residual connections, impeding proper data propagation.
For example, in the ViT model in Figure~\ref{fig:fig_4}, residual connections link attention and feed-forward layers. By default, they connect two layer outputs of the same dimension, as shown in Figure~\ref{fig:fig_4}-(a). However, after removing the neurons, the output dimension of the two layers is mismatched; see Figure~\ref{fig:fig_4}-(b). The solution in CNNSplitter~\cite{CNNSplitter} adds additional kernels to match the two residual connected layers in Figure~\ref{fig:fig_4}-(c), introducing additional memory and computational overhead. MwT~\cite{MwT} offers another solution by padding the output of each residual connected layer to match their dimension, shown in Figure~\ref{fig:fig_4}-(d). 
However, padding the back layer is redundant, as computations involving this padding yield zero. Only the padding of the front layer is valid since it needs to match the back layer for the next propagation. In Figure~\ref{fig:fig_4}-(e), \projectName offers the \textit{On-demand Padding} to apply padding only for the front layer and remove invalid calculations, avoiding additional overhead.

\begin{figure}[ht]
\centerline{\includegraphics[width=0.8\linewidth]{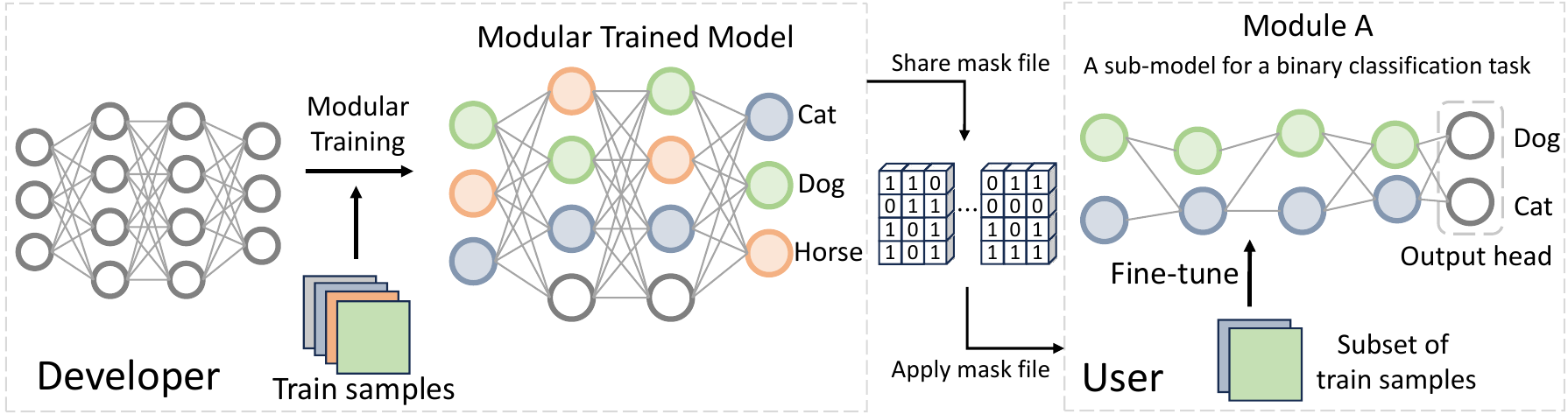}}
\caption{On-demand module reuse. }
\label{fig:fig_5}
\end{figure}

\subsection{On-demand reuse}

\projectName achieves efficient on-demand reuse through structural neuron removal, minimizing memory and computational overhead.  For instance, when tasked with identifying ``cat'' and ``dog'' categories, \projectName can extract the relevant module from a 10-class classification model, eliminating extraneous neurons and thereby reducing inference costs.

Specifically, in Figure~\ref{fig:fig_5}, after modular training, developers can apply structured modularization and release masks $M_{i}$ along with the modular trained model. 
Users can apply mask $M_{i}$ to the modular trained model and extract modules with specific functionalities. For composite modules with multiple functions, instead of combining multiple modules with each as independent component~\cite{CNNPan, DNNPan}, we apply $M_{D} = M_{B}\cup M_{C}$ to merge the masks of classes $B$ and $C$, retaining only one copy of neurons shared by both functions $A$ and $B$, thus significantly reducing overhead.
After decomposing the model to modules, considering the neurons of the original classification head are not modularized during the training process, a randomly initialized fully connected layer matching the module output and classification number is appended as the new classification head, mapping the features of module output to the target output. The resulting sub-model then undergoes fine-tuning for several epochs using a subset of the original training data specific to the target task. This process enables the sub-model to achieve accuracy comparable to the original model on the target task.

\section{Experiments}
We evaluate the effectiveness of \projectName by answering the following research questions:
\begin{itemize}
  \item RQ1: How effective is \projectName in training and modularizing DNN models?
  \item RQ2: How effective is \projectName in reusing DNN modules?
  \item RQ3: How effective is \projectName for CNN models compared with MwT?
  \item RQ4: How do the major hyper-parameters influence the performance of \projectName?
  \item RQ5: How does varying model scale affect \projectName's modularization efficiency compared to MwT?
\end{itemize}

\subsection{Experimental Setup}
\begin{table}[t]
\setlength\tabcolsep{3pt}
\caption{The settings of hyperparameters $\alpha$.}
\label{tab:alpha_hyperparameter}
\vspace{-6pt}
\centering
\resizebox{\columnwidth}{!}{
\begin{tabular}{ccccccccccccccc}
\toprule
& \multicolumn{3}{c}{\textbf{ViT}} & \multicolumn{3}{c}{\textbf{DeiT}} & \multicolumn{2}{c}{\textbf{VGG16}} & \multicolumn{2}{c}{\textbf{ResNet18}} & \multicolumn{2}{c}{\textbf{SimCNN}} & \multicolumn{2}{c}{\textbf{ResCNN}} \\ \cmidrule(lr){2-4} \cmidrule(lr){5-7} \cmidrule(lr){8-9} \cmidrule(lr){10-11} \cmidrule(lr){12-13} \cmidrule(lr){14-15}
& \textbf{CIFAR10}  & \textbf{SVHN} &\textbf{FashionMNIST}  & \textbf{CIFAR10}  & \textbf{SVHN} &\textbf{FashionMNIST}  & \textbf{CIFAR10}  & \textbf{SVHN}  & \textbf{CIFAR10} & \textbf{SVHN}& \textbf{CIFAR10}& \textbf{SVHN}  & \textbf{CIFAR10}& \textbf{SVHN}  \\ \midrule \midrule
\textbf{$\alpha$} & 0.2 & 0.1 & 0.2 & 0.1 & 0.1 & 0.2 & 1.4 & 1.6 & 1.2 & 1.2 & 1.3 & 1.0 & 1.0 & 1.0  
\\ \bottomrule
\end{tabular}
}
\end{table}

\textbf{Models.} (i) Vision Transformer (ViT)~\cite{ViT} and Data Efficient Image Transformer (DeiT)~\cite{deit} with $12$ encoders and each layer size of $384$.
(ii) Four representative CNN models, including ResNet18~\cite{residual}, VGG16~\cite{vgg}, SimCNN and ResCNN, which are also used by the baseline MwT~\cite{MwT}. 

\textbf{Datasets.} Three public classification datasets are used for standard and modular training, including CIFAR10, Street View House Number (SVHN), and FashionMINST, which are commonly used in DNN modularization works~\cite{CNNPan,MwT,SeaM,CNNSplitter}. 
The CIFAR10 dataset contains 50,000 natural images for training and 10,000 for testing with resolution $32\times 32$. Each sample includes a class from 10 classes: \textit{airplanes, cars, birds, cats, deer, dogs, frogs, horses, ships, } and \textit{trucks.} The SVHN dataset contains 604,388 house numbers from 0 to 9 for training and 26,032 for testing, with the resolution of $32\times 32$. The FashionMNIST contains 60,000 images for training and 10,000 for testing with the resolution $28\times 28$. Each sample includes a class from 10 classes: \textit{t-shirt, trouser, pullover, dress, coat, sandal, shirt, sneaker, bag, } and \textit{ankle boot.}

\textbf{Baselines.} (i) Standard training. Standard training optimizes ViT, DeiT, and CNN models using mini-batch stochastic gradient descent with cross-entropy loss. (ii) MwT~\cite{MwT}. MwT is the state-of-the-art modular training approach for CNN models. To compare \projectName and MwT on the ViT and DeiT models, we adapted MwT to the neuron level by replacing the kernel recognizer with a neuron identifier. 

\textbf{Evaluation Metrics.}
(i) Accuracy (ACC), which is calculated as the percentage of correct predictions on the whole test set. (ii) Neuron retention rate (NRR) and kernel retention rate (KRR), which indicate the average number of neurons or convolution kernels retained in the target module divided by the whole number of them in the original model. (iii) Cohesion, which is the average cohesion of all modules(Eq. (\ref{eq:cohesion}))~\cite{MwT}. (iv) Coupling, which is the average coupling across all pairs of modules (Eq.(\ref{eq:coupling}))~\cite{MwT}.

\textbf{Hyper-parameters.} In standard training, ViT, DeiT, ResNet18, SimCNN, and ResCNN are trained for 200 epochs using a mini-batch size of 128, and VGG16 is trained with a mini-batch size of 256. We set the learning rate to 0.05 and Nesterov’s momentum to 0.9. As for data augmentation\cite{data_aug}, we apply random cropping and flipping for all the models. The settings of contrastive loss weight $\alpha$ are shown in Table ~\ref{tab:alpha_hyperparameter}, the temperature $\tau$ in the contrastive loss for all models are set to $0.2$.

\textbf{Experiment phases.} For \projectName, it consists of two phases. The first phase is modular training, where \projectName decouples neurons during training by adding neuron identifiers to corresponding linear layers, including MLP and attention layers. The second phase is decomposition, where \projectName performs a forward pass with the training set to get masks that represent relationships between neurons and corresponding classes at each layer. In these masks, 0 indicates that a neuron is unrelated to the target class, while non-zero values indicate relevance. Using these masks, \projectName decomposes the model into all modules (for example, in a CIFAR-10 classification model, 10 modules are obtained) and fine-tunes each module on its target class task for several epochs. MwT follows a similar process. In RQ1, to demonstrate that modularized training does not significantly reduce performance, we designed a standard training baseline where the model is trained without any modularization methods to serve as an accuracy baseline.

\subsection{Experimental Results}

\begin{table}[t]
\caption{The comparison of \projectName and standard training on the Transformer-based models.}
\label{tab:rq1_st_and_mt_vit}
\vspace{-6pt}
\centering
\resizebox{0.7\columnwidth}{!}{
\begin{tabular}{c c c c c c c c}
\toprule
\multirow{2}{*}{\textbf{Model}} & \multirow{2}{*}{\textbf{Dataset}} & \multirow{2}{*}{\textbf{\#Neurons}} & \multirow{2}{*}{\textbf{\begin{tabular}[c]{@{}c@{}}Standard\\  Model ACC\end{tabular}}} & \multirow{2}{*}{\textbf{\begin{tabular}[c]{@{}c@{}}Modular\\ Model ACC\end{tabular}}} & \multicolumn{3}{c}{\textbf{Modules}}       
\\ \cmidrule(lr){6-8} &    &    &    &    & \textbf{NRR} & \textbf{Cohesion} & \textbf{Coupling} 
\\ \midrule \midrule
\multirow{3}{*}{ViT} 
& CIFAR10  & \multirow{3}{*}{41472}  & 77.20  & 76.57  & 8.03  & 0.9061  & 0.0812 
\\  & SVHN  &  & 94.34  & 92.78  & 9.21  & 0.9096  & 0.2230
\\  & FashionMNIST  &  & 92.61  & 91.58  & 12.04  & 0.9283  & 0.0899 
\\ \midrule
\multirow{3}{*}{DeiT} 
& CIFAR10  & \multirow{3}{*}{41472}  & 78.40  & 78.26  & 15.02  & 0.9688  & 0.1392 
\\  & SVHN  &  & 95.29  & 91.70  & 11.92  & 0.9262  & 0.1992
\\  & FashionMNIST  &  & 91.75  & 91.32  & 13.32  & 0.9372  & 0.1134
\\ \midrule
\multicolumn{2}{c}{\textbf{Average}} 
& \textbf{41472}    & \textbf{88.27}    & \textbf{87.04}    & \textbf{11.59} & \textbf{0.9294}   & \textbf{0.1410}  
\\ \bottomrule
\end{tabular}
    }
\end{table}

\subsubsection{RQ1 - Effectiveness of Modular Training and Modularizing}
To evaluate the effectiveness of \projectName in training and modularizing models, we apply \projectName on two Transformer-based models. Furthermore, we replace the \textit{relevant kernel recognition} in MwT with \textit{neuron identifier} to make MwT feasible in the Transformer-based models for comparison. 
We evaluate (1) the classification accuracy of modular models, (2) the neuron retention rate (NRR) for Transformer-based models, and (3) the cohesion and coupling of trained modules.
Table ~\ref{tab:rq1_st_and_mt_vit} shows the results of standard training, modular training, and modularizing on Transformer-based models. The ``\#Neurons'' column shows the number of neurons for the ViT and DeiT models. 
The ``Standard Model ACC'' and ``Modular Model ACC'' show the accuracy of the standard training model and modular training model on the test set, respectively. 
For the Transformer-based models ViT and DeiT, the standard and modular models achieve an accuracy of 88.27\% and 87.04\%, indicating a loss of only 1.23 percentage points for modular training, showing that modular training does not cause much accuracy loss for classification. 
As for the Neuron Retention Rate (NRR), the average NRR for the ViT and DeiT models is 11.59\%, which means after modularization,  with the threshold of 0.9, each module retained 11.59\% neurons on average. 
The cohesion degrees for Transformer-based models are 0.9294, indicating that a module uses almost the same neurons to predict the samples belonging to the corresponding class. 
On the other hand, the coupling degree for Transformer-based models is 0.1410, demonstrating that different modules share only a few neurons.

Moreover, we also compare \projectName with MwT, the state-of-the-art modular training approach, for the ViT and DeiT models, as shown in Table ~\ref{tab:rq1_vit_mwt_ours}.
Similar to CNN models in MwT, which apply masks on the output channel of convolution layers to represent which convolution kernel should be selected and generated to a module, the Transformer-based models can use masks on the output of the linear layer to represent which neurons should be selected. Based on these masks, we can apply MwT to the Transformer-based models and use the masks to decompose the model structurally.
On average, the accuracy of the modular trained model for MwT and \projectName is 80.03\% and 87.04\%, respectively. MwT causes an accuracy loss of 6.28 percentage points, while \projectName causes an accuracy loss of only 1.48 percentage points. 
On the other hand, \projectName also outperforms MwT in terms of NRR (11.59\% vs 26.37\%), with a reduction of 58.10\%.
\begin{table}[t]
\caption{The comparison of \projectName and MwT on Transformer-based models.}
\label{tab:rq1_vit_mwt_ours}
\vspace{-6pt}
\centering
\resizebox{0.8\columnwidth}{!}{
\begin{tabular}{ccccccccccc}
\toprule
\multirow{2}{*}{\textbf{Model}}  & \multirow{2}{*}{\textbf{Dataset}} & \multirow{2}{*}{\textbf{\begin{tabular}[c]{@{}c@{}}Standard \\ Model ACC\end{tabular}}} 
& \multicolumn{4}{c}{\textbf{MwT}} & \multicolumn{4}{c}{\textbf{\projectName}}  
\\ 
\cmidrule(lr){4-7} \cmidrule(lr){8-11}  
&  &  & \textbf{ACC}   & \textbf{Coupling} & \textbf{Cohesion} & \textbf{NRR}& \textbf{ACC}  & \textbf{Coupling} & \textbf{Cohesion} & \textbf{NRR}
\\ \midrule \midrule
\multirow{2}{*}{ViT} & CIFAR10   & 77.20   & 75.19  & 0.2205  & 0.9224  & 20.53& 76.57  & 0.0812  & 0.9061  & 8.03
\\ \cmidrule(lr){2-11} & SVHN  & 94.34  & 92.89  & 0.2648  & 0.9421  & 28.38  & 92.78  & 0.2230  & 0.9096  & 9.21
\\ \midrule 
\multirow{2}{*}{DeiT} & CIFAR10   & 78.40   & 73.31  & 0.2783  & 0.9508  & 31.21  & 78.26  & 0.1392  & 0.9688  & 15.02
\\ \cmidrule(lr){2-11} & SVHN  & 95.29   & 78.74  & 0.2477  & 0.8895  & 25.37  & 91.70  & 0.1992  & 0.9262  & 11.92
\\ \midrule 
\multicolumn{2}{c}{\textbf{Average}}  & \textbf{86.31}  & \textbf{80.03} & \textbf{0.2453}   & \textbf{0.9262}  & \textbf{26.37}  & \textbf{84.83} & \textbf{0.1606}  & \textbf{0.9277} & \textbf{11.05}
\\ \bottomrule
\end{tabular}
}
\end{table}
Overall, compared to MwT, Transformer-based models trained with \projectName exhibit higher cohesion and lower coupling, and at the same time, higher accuracy and lower NRR.

\begin{tcolorbox}[colback=white, colframe=black, boxrule=0.4mm, arc=2mm, left=3mm, right=3mm, top=1mm, bottom=1mm, width=\columnwidth]
\projectName outperforms the state-of-the-art approach MwT in both modular training and modularizing on all models. Compared to MwT, \projectName achieves higher accuracy (with an improvement of 4.8 percentage points) while gains average improvements of 5.42, 1.33, and 15.32 percentage points in coupling, cohesion, and NRR respectively.
\end{tcolorbox}

\subsubsection{RQ2 - Module Reuse Effectiveness}
In this RQ, we investigate the effectiveness of \projectName in on-demand model reuse and compare \projectName with MwT. Specifically, for the two 10-class classification tasks corresponding to the CIFAR10 and SVHN datasets, each task can be divided into n-class classification sub-tasks with the number of categories $n$ ranging from 2 to 10. Considering the huge number of sub-tasks (e.g., there are $C_{10}^{3}$ 3-class classification sub-tasks), we randomly select 10 sub-tasks for each n-class classification scenario.
In each n-class classification scenario, we reuse the module on demand from \projectName and MwT, then analyze the average number of neurons and weights within the models, as well as the average accuracy of the models.
\begin{table}[t]
\setlength\tabcolsep{3pt}
\caption{The comparison of \projectName and MwT in reusing the ViT modules in terms of Neuron Retention Rate(NRR) and Weight Retention Rate(WRR). All results in \%.}
\label{tab:rq2_mwt_and_ours_nrr}
\vspace{-6pt}
\centering
\begin{minipage}[t]{0.48\linewidth}
\centering
\caption*{(a) Neuron Retention Rate (NRR)}
\resizebox{\columnwidth}{!}{
\begin{tabular}{ccccccc}
\toprule
\multirow{2}{*}{\textbf{\begin{tabular}[c]{@{}c@{}}Target\\Task\end{tabular}}} & \multirow{2}{*}{\textbf{Approach}} 
& \multicolumn{2}{c}{\textbf{ViT}} & \multicolumn{2}{c}{\textbf{DeiT}} & \multirow{2}{*}{\textbf{Average}} 
\\
\cmidrule(lr){3-4} \cmidrule(lr){5-6} &  
& \textbf{CIFAR10}   & \textbf{SVHN} 
& \textbf{CIFAR10}   & \textbf{SVHN} 
\\
\midrule \midrule
\multirow{2}{*}{2-class} 
& \projectName   & 18.80  & 20.22 & 28.99 & 21.95 & 22.49 \\
& MwT  & 44.48 & 44.00 & 44.03 & 44.43 & 44.24 \\ 
\hline
\multirow{2}{*}{3-class} 
& \projectName  & 22.58  & 28.92 & 36.26 & 31.61 & 29.84 \\
& MwT  & 49.08 & 65.67 & 60.30 & 80.40 & 63.86 \\ 
\hline
\multirow{2}{*}{4-class} 
& \projectName  & 25.11  & 31.07 & 42.78 & 37.63 & 34.15 \\
& MwT  & 62.54 & 87.76 & 83.56 & 80.80 & 78.67 \\ 
\hline
\multirow{2}{*}{5-class} 
& \projectName  & 29.95  & 37.46 & 50.73 & 44.41 & 40.64 \\
& MwT  & 76.21 & 88.00 & 85.62 & 80.85 & 82.67 \\ 
\hline
\multirow{2}{*}{6-class} 
& \projectName  & 31.67  & 40.62 & 57.46 & 48.42 & 44.54 \\
& MwT  & 76.77 & 88.50 & 86.41 & 82.15 & 83.46 \\ 
\hline
\multirow{2}{*}{7-class} 
& \projectName  & 39.12  & 41.26 & 63.87 & 50.53 & 48.70 \\
& MwT  & 83.47 & 88.56 & 88.17 & 82.16 & 85.59 \\ 
\hline
\multirow{2}{*}{8-class} 
& \projectName  & 44.45  & 44.80 & 70.14 & 54.14 & 53.38 \\
& MwT  & 83.89 & 88.62 & 88.78 & 82.19 & 85.87 \\ 
\hline
\multirow{2}{*}{9-class} 
& \projectName  & 52.99  & 45.44 & 77.19 & 56.24 & 57.97 \\
& MwT  & 84.38 & 88.64 & 89.51 & 82.24 & 86.19 \\ 
\hline
\multirow{2}{*}{10-class}
& \projectName  & 57.96  & 46.15 & 81.10 & 57.94 & 60.79 \\
& MwT  & 84.49 & 88.88 & 89.69 & 82.25 & 86.33 \\ 
\bottomrule
\end{tabular}
}
\end{minipage}
\hspace{0.02\linewidth}
\begin{minipage}[t]{0.48\linewidth}
\centering
\caption*{(b) Weight Retention Rate (WRR)}
\resizebox{\columnwidth}{!}{
\begin{tabular}{ccccccc}
\toprule
\multirow{2}{*}{\textbf{\begin{tabular}[c]{@{}c@{}}Target\\Task\end{tabular}}} & \multirow{2}{*}{\textbf{Approach}} 
& \multicolumn{2}{c}{\textbf{ViT}} & \multicolumn{2}{c}{\textbf{DeiT}} & \multirow{2}{*}{\textbf{Average}} 
\\
\cmidrule(lr){3-4} \cmidrule(lr){5-6} &  
& \textbf{CIFAR10}   & \textbf{SVHN} 
& \textbf{CIFAR10}   & \textbf{SVHN} 
\\
\midrule \midrule
\multirow{2}{*}{2-class} 
& \projectName   & 12.04  & 13.71 & 17.00 & 14.81 & 14.39 \\
& MwT  & 25.15 & 24.16 & 25.05 & 25.56 & 24.98 \\ 
\hline
\multirow{2}{*}{3-class} 
& \projectName  & 14.26  & 19.08 & 22.05 & 21.94 & 19.33 \\
& MwT  & 29.45 & 44.87 & 41.03 & 71.07 & 46.61 \\ 
\hline
\multirow{2}{*}{4-class} 
& \projectName  & 16.24  & 20.88 & 27.69 & 26.86 & 22.92 \\
& MwT  & 43.04 & 76.60 & 75.07 & 71.46 & 66.54 \\ 
\hline
\multirow{2}{*}{5-class} 
& \projectName  & 19.51  & 26.09 & 35.57 & 33.93 & 28.78 \\
& MwT  & 61.09 & 77.04 & 77.11 & 71.51 & 71.69 \\ 
\hline
\multirow{2}{*}{6-class} 
& \projectName  & 20.98  & 29.20 & 43.28 & 39.02 & 33.12 \\
& MwT  & 61.66 & 77.65 & 77.83 & 72.88 & 72.51 \\ 
\hline
\multirow{2}{*}{7-class} 
& \projectName  & 27.09  & 29.88 & 50.73 & 41.62 & 37.33 \\
& MwT  & 71.04 & 77.76 & 79.55 & 72.90 & 75.31 \\ 
\hline
\multirow{2}{*}{8-class} 
& \projectName  & 32.35  & 33.27 & 58.91 & 45.90 & 42.61 \\
& MwT  & 71.54 & 77.84 & 80.27 & 72.94 & 75.65 \\ 
\hline
\multirow{2}{*}{9-class} 
& \projectName  & 41.42  & 33.96 & 68.49 & 48.64 & 48.13 \\
& MwT  & 72.16 & 77.87 & 81.09 & 72.99 & 76.03 \\ 
\hline
\multirow{2}{*}{10-class}
& \projectName  & 47.31  & 34.65 & 73.89 & 50.69 & 51.64 \\
& MwT  & 72.29 & 78.25 & 81.31 & 73.00 & 76.21 \\ 
\bottomrule
\end{tabular}
}
\end{minipage}
\end{table}

Table ~\ref{tab:rq2_mwt_and_ours_nrr} presents the neuron retention and weight retention rates on different classification sub-tasks for the ViT model. For instance, in the left table of Table ~\ref{tab:rq2_mwt_and_ours_nrr}, for a 2-class CIFAR10 classification sub-task, the corresponding ViT-CIFAR10 module from \projectName utilizes only 18.80\% of the model's neurons, while the same module from MwT has 44.48\% of the model's neurons. On average, the overall neuron retention rate for \projectName has a 43.68\% reduction (43.61\% vs 77.34\%, simply calculate the average value for all modules) than that of MwT, significantly reducing the module size and operational overhead.
Notably, the ViT and DeiT models employed in our study consist of 12 encoders. To ensure the accuracy of the modules, we retained the complete first encoder and decomposed the subsequent 11 encoders. Consequently, all NRR values represent the neuron retention rates in the latter 11 encoders. To more accurately evaluate the number of parameters retained by the module, we used WRR to quantify the weight retention rates across all 12 encoders. In the right table of Table ~\ref{tab:rq2_mwt_and_ours_nrr}, we considered all encoders in ViT and DeiT models and calculated the weight retention rates. For those models, weight retention rates more directly reflect the situation of parameters compared to neuron retention rates. On average, the modules from \projectName retained only 33.14\% weights of the original model, a 49.06\% reduction than the module from MwT, which has 65.06\% weights of the model.
For each sub-task, \projectName is capable of generating smaller modules compared to MwT.

\begin{table}[t]
\caption{The test accuracy results of \projectName in reusing ViT and DeiT modules. All results in \%.}
\label{tab:rq2_standard_and_ours_acc}
\vspace{-6pt}
\centering
\resizebox{0.5\columnwidth}{!}{
\begin{tabular}{ccrrrrr}
\toprule
\multirow{2}{*}{\textbf{\begin{tabular}[c]{@{}c@{}}Target\\Task\end{tabular}}} & \multirow{2}{*}{\textbf{Approach}} & \multicolumn{2}{c}{\textbf{ViT}} & \multicolumn{2}{c}{\textbf{DeiT}} & \multirow{2}{*}{\textbf{Average}}
\\ \cmidrule(lr){3-4} \cmidrule(lr){5-6} 
&  & \textbf{CIFAR10}  & \textbf{SVHN}  & \textbf{CIFAR10}  & \textbf{SVHN}  &  
\\ \midrule \midrule
\multirow{3}{*}{2-class}  
& MwT  & 95.10  & 98.96  & 95.15  & 96.67  & 96.47  \\  
& \projectName  & 94.55  & 98.32  & 95.40  & 98.23  & 96.63  
\\ \cline{2-7}
& \textbf{loss}  & 0.55  & 0.64  & \cellcolor{gray!30}-0.25  & \cellcolor{gray!30}-1.56  & \cellcolor{gray!30}-0.16  
\\ \hline
\multirow{3}{*}{3-class}  
& MwT  & 90.60  & 98.31  & 90.90  & 95.34  & 93.79  \\
& \projectName  & 91.43  & 97.21  & 92.17  & 96.81  & 94.41
\\ \cline{2-7}
& \textbf{loss}  & \cellcolor{gray!30}-0.83  & 1.10  & \cellcolor{gray!30}-1.27  & \cellcolor{gray!30}-1.47  & \cellcolor{gray!30}-0.62  
\\ \hline
\multirow{3}{*}{4-class}  
& MwT  & 85.55  & 96.54  & 86.35  & 90.87  & 89.83  \\
& \projectName  & 87.45  & 95.78  & 86.40  & 95.12  & 91.19 
\\ \cline{2-7}
& \textbf{loss}  & \cellcolor{gray!30}-1.90  & 0.76  & \cellcolor{gray!30}-0.05  & \cellcolor{gray!30}-4.25  & \cellcolor{gray!30}-1.36 
\\ \hline
\multirow{3}{*}{5-class}  
& MwT  & 81.70  & 95.62  & 83.24  & 90.10  & 87.67  \\
& \projectName  & 83.46  & 95.08  & 83.24  & 94.09  & 88.97
\\ \cline{2-7}
& \textbf{loss}  & \cellcolor{gray!30}-1.76  & 0.54  & 0.00  & \cellcolor{gray!30}-3.99  & \cellcolor{gray!30}-1.30
\\ \hline
\multirow{3}{*}{6-class}  
& MwT  & 76.28  & 94.62  & 77.30  & 86.26  & 83.62  \\
& \projectName  & 77.75  & 94.18  & 80.90  & 92.99  & 86.46
\\ \cline{2-7}
& \textbf{loss}  & \cellcolor{gray!30}-1.47  & 0.44  & \cellcolor{gray!30}-3.60  & \cellcolor{gray!30}-6.73  & \cellcolor{gray!30}-2.84 
\\ \hline
\multirow{3}{*}{7-class}  
& MwT  & 76.30  & 94.12  & 77.13  & 85.02  & 83.41 \\
& \projectName  & 77.14  & 93.09  & 79.93  & 92.18  & 85.59 
\\ \cline{2-7}
& \textbf{loss}  & \cellcolor{gray!30}-0.84  & 1.03  & \cellcolor{gray!30}-2.80  & \cellcolor{gray!30}-7.16  & \cellcolor{gray!30}-2.18
\\ \hline
\multirow{3}{*}{8-class}  
& MwT  & 75.55  & 93.81  & 76.13  & 84.75  & 82.56 \\
& \projectName  & 77.01  & 92.94  & 78.76  & 91.79  & 85.13
\\ \cline{2-7}
& \textbf{loss}  & \cellcolor{gray!30}-1.46  & 1.03  & \cellcolor{gray!30}-2.63  & \cellcolor{gray!30}-7.04  & \cellcolor{gray!30}-2.57
\\ \hline
\multirow{3}{*}{9-class}  
& MwT  & 75.90  & 92.64  & 75.81  & 82.50  & 81.71  \\
& \projectName  & 77.17  & 92.26  & 80.04  & 87.40  & 84.22  
\\ \cline{2-7}
& \textbf{loss}  & \cellcolor{gray!30}-1.27  & 0.38  & \cellcolor{gray!30}-4.23  & \cellcolor{gray!30}-4.90  & \cellcolor{gray!30}-2.51
\\ \hline
\multirow{3}{*}{10-class}  
& MwT  & 74.32  & 92.71  & 73.79  & 83.06  & 80.97  \\
& \projectName  & 76.15  & 91.61  & 79.31  & 90.45  & 84.38 
\\ \cline{2-7}
& \textbf{loss}  & \cellcolor{gray!30}-1.83  & 1.10  & \cellcolor{gray!30}-5.52  & \cellcolor{gray!30}-7.39  & \cellcolor{gray!30}-3.41
\\ \bottomrule
\end{tabular}
}
\end{table}
We further compare \projectName with MwT in terms of the accuracy of on-demand model reuse in Table ~\ref{tab:rq2_standard_and_ours_acc}. The ``loss'' row in the table represents the extent to which \projectName loses accuracy compared to MwT. A positive value indicates that \projectName's accuracy is lower than that of MwT, while a negative value indicates that \projectName's accuracy is higher than MwT. 
Essentially, across all models and sub-tasks, \projectName maintains the same accuracy as MwT. Considering that the modules generated by \projectName for all sub-tasks retain almost half the number of neurons compared to MwT, \projectName's performance in terms of accuracy is commendable.

To evaluate the extent to which \projectName simplifies on-demand model reuse, we measured the FLOPs(M) of all modules on target tasks and compared \projectName's results with both MwT and standard trained models in Table~\ref{tab:rq2_flops}. The values in the model name row represent the FLOPs of standard trained models when directly reused. Our experiments demonstrate that \projectName significantly simplifies on-demand model reuse, reducing FLOPs by up to 82.99\%. Compared to MwT, \projectName also shows a substantial improvement in simplifying on-demand reuse, with FLOPs decreasing by up to 66.73\%.

\begin{table}[t]
\setlength\tabcolsep{3pt}
\newcommand{\reductionpercentages}[2]{%
  \raisebox{-0.5ex}{\scriptsize\textcolor{blue}{$\downarrow$#1\%},\textcolor{red}{$\downarrow$#2\%}}%
}
\caption{The comparison of \projectName and MwT for FLOPs in reusing the ViT modules, all results in FLOPs (M).}
\label{tab:rq2_flops}
\vspace{-6pt}
\centering
\resizebox{0.65\columnwidth}{!}{
\begin{tabular}{c c c c c c c}
\toprule
\multirow{2}{*}{\textbf{\begin{tabular}[c]{@{}c@{}}Target\\Task\end{tabular}}} & \multirow{2}{*}{\textbf{Approach}} 
& \multicolumn{2}{c}{\textbf{ViT(1384.49)}} & \multicolumn{2}{c}{\textbf{DeiT(1405.78)}} & \multirow{2}{*}{\textbf{Average(1395.14)}} 
\\
\cmidrule(lr){3-4} \cmidrule(lr){5-6} &  
& \textbf{CIFAR10}   & \textbf{SVHN} 
& \textbf{CIFAR10}   & \textbf{SVHN} 
\\
\midrule \midrule
\multirow{2}{*}{2-class} 
& \projectName   & 198.56  & 226.21 & 277.23 & 246.83 & 237.21\reductionpercentages{38.43}{82.99} \\
& MwT  & 382.17  & 371.23 & 390.25 & 397.45 & 385.28 \\ 
\hline
\multirow{2}{*}{3-class} 
& \projectName   & 229.64  & 301.08 & 348.27 & 347.19 & 306.55\reductionpercentages{59.71}{78.03} \\
& MwT  & 735.03  & 657.60 & 614.73 & 1036.01 & 760.84 \\ 
\hline
\multirow{2}{*}{4-class} 
& \projectName   & 257.36  & 326.35 & 427.61 & 416.11 & 356.86\reductionpercentages{66.73}{74.42} \\
& MwT  & 1058.40  & 1096.45 & 1092.90 & 1042.29 & 1072.51 \\  
\hline
\multirow{2}{*}{5-class} 
& \projectName   & 303.14  & 399.11 & 538.32 & 515.75 & 439.08\reductionpercentages{59.46}{68.53} \\
& MwT  & 1065.04  & 1102.59 & 1121.46 & 1043.04 & 1083.03 \\ 
\hline
\multirow{2}{*}{6-class} 
& \projectName   & 323.72  & 442.24 & 646.66 & 587.16 & 499.95\reductionpercentages{54.29}{64.16} \\
& MwT  & 1070.52  & 1111.13 & 1131.55 & 1062.19 & 1093.85 \\ 
\hline
\multirow{2}{*}{7-class} 
& \projectName   & 409.26  & 452.23 & 751.15 & 623.75 & 559.10\reductionpercentages{49.35}{59.93} \\
& MwT  & 1084.95  & 1112.63 & 1155.66 & 1062.40 & 1103.91 \\ 
\hline
\multirow{2}{*}{8-class} 
& \projectName   & 482.90  & 499.33 & 866.13 & 683.77 & 633.03\reductionpercentages{43.16}{54.63} \\
& MwT  & 1113.07  & 1113.66 & 1165.69 & 1062.95 & 1113.84 \\ 
\hline
\multirow{2}{*}{9-class} 
& \projectName   & 609.88  & 508.57 & 1000.54 & 722.23 & 710.31\reductionpercentages{36.54}{49.09} \\
& MwT  & 1121.94  & 1114.06 & 1177.18 & 1063.69 & 1119.22 \\ 
\hline
\multirow{2}{*}{10-class}
& \projectName   & 692.34  & 518.07 & 1076.28 & 751.02 & 759.43\reductionpercentages{32.29}{45.57} \\
& MwT  & 1123.00  & 1119.40 & 1180.35 & 1063.81 & 1121.64 \\ 
\bottomrule
\end{tabular}
}
\end{table}

\begin{tcolorbox}[colback=white, colframe=black, boxrule=0.4mm, arc=2mm, left=3mm, right=3mm, top=1mm, bottom=1mm, width=\columnwidth]
Modules generated by \projectName can achieve up to a 58.10\% reduction in the number of neurons and 66.73\% reduction in FLOPs compared to MwT through on-demand reuse, with virtually no loss in accuracy.

\end{tcolorbox}

\begin{table}[t]
\caption{The comparison of \projectName and standard training on CNN models.}
\label{tab:rq1_st_and_mt}
\vspace{-6pt}
\centering
\resizebox{0.7\columnwidth}{!}{
\begin{tabular}{cccccccc}
\toprule
\multirow{2}{*}{\textbf{Model}} & \multirow{2}{*}{\textbf{Dataset}} & \multirow{2}{*}{\textbf{\#Kernels}} & \multirow{2}{*}{\textbf{\begin{tabular}[c]{@{}c@{}}Standard\\  Model ACC\end{tabular}}} & \multirow{2}{*}{\textbf{\begin{tabular}[c]{@{}c@{}}Modular\\ Model ACC\end{tabular}}} & \multicolumn{3}{c}{\textbf{Modules}}       
\\ \cmidrule(lr){6-8} &    &    &    &    & \textbf{KRR} & \textbf{Cohesion} & \textbf{Coupling} 
\\ \midrule \midrule
\multirow{2}{*}{VGG16} 
& CIFAR10  & \multirow{2}{*}{4224}  & 92.29  & 90.15  & 14.25  & 0.9784  & 0.1959 
\\  & SVHN  &  & 95.84  & 94.99  & 10.96  & 0.9703  & 0.1551 
\\ \midrule
\multirow{2}{*}{ResNet18} 
& CIFAR10   & \multirow{2}{*}{3904}    & 92.29  & 90.39  & 15.64   & 0.9802   & 0.2103
\\  & SVHN  &  & 95.84  & 95.34  & 14.61  & 0.9665  & 0.2136 
\\ \midrule
\multirow{2}{*}{SimCNN}
& CIFAR10   & \multirow{2}{*}{4224}    & 89.77  & 89.84  & 11.37   & 0.9681    & 0.2517
\\  & SVHN  &  & 95.41  & 95.19  & 10.47   & 0.9751   & 0.1185
\\ \midrule
\multirow{2}{*}{ResCNN} 
& CIFAR10   & \multirow{2}{*}{4288}    & 90.41  & 90.39  & 15.38   & 0.9640   & 0.2083
\\  & SVHN  &  & 95.06  & 94.40  & 9.69   & 0.9651   & 0.2066
\\ \midrule
\multicolumn{2}{c}{\textbf{Average}} 
& \textbf{4160}    & \textbf{93.50}    & \textbf{92.59}    & \textbf{12.80} & \textbf{0.9710}   & \textbf{0.1950}   
\\ \bottomrule
\end{tabular}
}
\end{table}

\begin{table}[t]
\caption{The comparison of  \projectName and MwT on CNN models.}
\label{tab:rq1_cnn_mwt_ours}
\vspace{-6pt}
\centering
\resizebox{0.9\columnwidth}{!}{
\begin{tabular}{cccccccccccc}
\toprule
\multirow{2}{*}{\textbf{Model}}  & \multirow{2}{*}{\textbf{Dataset}} & \multirow{2}{*}{\textbf{\#Kernels}} & \multirow{2}{*}{\textbf{\begin{tabular}[c]{@{}c@{}}Standard \\ Model ACC\end{tabular}}} 
& \multicolumn{4}{c}{\textbf{MwT}} & \multicolumn{4}{c}{\textbf{\projectName}}          
\\ \cmidrule(lr){5-8} \cmidrule(lr){9-12}  
& &  &  & \textbf{ACC}   & \textbf{Coupling} & \textbf{Cohesion} & \textbf{KRR} & \textbf{ACC}  & \textbf{Coupling} & \textbf{Cohesion} & \textbf{KRR} \\ \midrule \midrule
\multirow{2}{*}{VGG16}  & CIFAR10  & \multirow{2}{*}{4224}  & 92.29   & 90.86  & 0.1751  & 0.9758  & 17.28  & 90.15  & 0.1959  & 0.9784  & 14.25 
\\ & SVHN &  & 95.84  & 94.74  & 0.2246  & 0.9687  & 14.15  & 94.99  & 0.1551  & 0.9703  & 10.96 
\\ \midrule 
\multirow{2}{*}{ResNet18} & CIFAR10 & \multirow{2}{*}{3904}  & 93.39   & 91.59  & 0.2412  & 0.9437  & 24.74  & 90.39   & 0.2103   & 0.9802  & 15.64 
\\ & SVHN &   & 95.84  & 95.95  & 0.3115  & 0.9663  & 25.89  & 95.34  & 0.2136  & 0.9665  & 14.61
\\ \midrule 
\multirow{2}{*}{SimCNN} & CIFAR10 & \multirow{2}{*}{4224}  & 89.77   & 88.84  & 0.1372  & 0.8682  & 11.58  & 89.84   & 0.2517   & 0.9681  & 11.37 
\\ & SVHN &   & 95.41   & 93.56  & 0.1434  & 0.9580  & 11.85  & 95.19   & 0.1185   & 0.9751  & 10.47
\\ \midrule 
\multirow{2}{*}{ResCNN} & CIFAR10 & \multirow{2}{*}{4288}  & 90.41   & 89.82  & 0.2781  & 0.9601  & 21.52  & 90.39   & 0.2083   & 0.9640  & 15.38
\\ & SVHN &   & 95.06   & 93.88  & 0.3306  & 0.9731  & 13.37  & 94.40   & 0.2066   & 0.9651  & 9.69
\\ \midrule 
\multicolumn{2}{c}{\textbf{Average}}  & \textbf{4160}  & \textbf{93.50}  & \textbf{92.41} & \textbf{0.2302}   & \textbf{0.9518}  & \textbf{17.55}  & \textbf{92.59} & \textbf{0.1950}  & \textbf{0.9710} & \textbf{12.80} 
\\ \bottomrule
\end{tabular}
}
\end{table}

\subsubsection{RQ3 - Comparison with MwT on CNNs}
To assess the generalizability of \projectName,
we conducted a comprehensive comparative analysis between \projectName and MwT across four CNN models. This evaluation serves two primary purposes: (1) To demonstrate the efficacy of contrastive learning in optimizing cohesion and coupling loss. (2) To establish that \projectName's effectiveness extends beyond Transformer-based architectures, showing its generalizability across diverse neural network structures. 
To ensure a fair comparison, we used the same four CNN models as in the original MwT study. All MwT data were sourced directly from the original paper or its source code. This approach guarantees consistency in model architectures and data, enabling an unbiased assessment of \projectName's performance against MwT across various CNN models. Specifically, we evaluated the effectiveness of \projectName in modular training and on-demand reuse.

\textbf{Modular Training.}
To evaluate the effectiveness of \projectName in training CNN models, we replace the \textit{neuron identifier} with \textit{relevant kernel recognition} ~\cite{MwT}, calculating cohesion and coupling with convolution kernel masks to make \projectName applicable. We evaluate (1) the classification accuracy of modular models, (2) the kernel retention rate (KRR) for four CNN models, and (3) the cohesion and coupling of trained modules. Table ~\ref{tab:rq1_st_and_mt} shows the results of standard training, modular training, and modularizing on CNN models. The ``\#Kernels'' column indicates the number of convolution kernels for different CNN models. The ``Standard Model ACC'' and ``Modular Model ACC'' show the accuracy of the standard training model and modular training model on the test set, respectively. 
The standard and modular models achieve an accuracy of 93.50\% and 92.59\%, and the average accuracy loss for modular training is 0.91 percentage points, showing that the modular training process of \projectName does not cause much accuracy loss for classification. 
In Table ~\ref{tab:rq1_cnn_mwt_ours}, we compare \projectName with MwT.
The accuracy of the modular trained model for MwT and \projectName are 92.41\% and 92.59\%, respectively, indicating an improvement of 0.18 percentage points achieved by \projectName. 
Also, \projectName achieves a decent performance gain over MwT (17.55\% vs 12.80\%) in terms of KRR, with an improvement of 27.07\%.
Regarding cohesion and coupling, we employed the MwT evaluation algorithm and assessed the cohesion and coupling of models trained using \projectName. \projectName achieves 0.9710 in cohesion and 0.1950 in coupling, which are 0.9518 and 0.2302 for MwT. Overall, compared to MwT, models trained with \projectName exhibit higher cohesion and lower coupling, and at the same time, higher accuracy and lower NRR/KRR.

\begin{table}[t]
\setlength\tabcolsep{3pt}
\caption{The comparison of \projectName and MwT in reusing CNN modules in terms of KRR. All results in \%.}
\label{tab:rq2_mwt_and_ours_krr}
\vspace{-6pt}
\centering
\resizebox{0.8\columnwidth}{!}{
\begin{tabular}{ccccccccccc}
\toprule
\multirow{2}{*}{\textbf{Target Task}} & \multirow{2}{*}{\textbf{Approach}} 
& \multicolumn{2}{c}{\textbf{VGG16}}  & \multicolumn{2}{c}{\textbf{ResNet18}}  & \multicolumn{2}{c}{\textbf{SimCNN}}  & \multicolumn{2}{c}{\textbf{ResCNN}}  & \multirow{2}{*}{\textbf{Average}} 
\\ 
\cmidrule(lr){3-4} \cmidrule(lr){5-6} \cmidrule(lr){7-8} \cmidrule(lr){9-10} &  
& \textbf{CIFAR10}   & \textbf{SVHN}  & \textbf{CIFAR10}   & \textbf{SVHN}  & \textbf{CIFAR10}   & \textbf{SVHN}  & \textbf{CIFAR10}   & \textbf{SVHN}  &
\\ \midrule \midrule
\multirow{2}{*}{2-class} 
& \projectName   & \cellcolor{gray!30}24.78  & \cellcolor{gray!30}22.24  & \cellcolor{gray!30}25.78  & \cellcolor{gray!30}26.29  & \cellcolor{gray!30}19.52  & \cellcolor{gray!30}22.88  & \cellcolor{gray!30}24.69  & 21.16  & \cellcolor{gray!30}23.42 \\
& MwT  & 30.34 & 27.38 & 35.94 & 42.44  & 20.34  & 34.13  & 28.69  & \cellcolor{gray!30}19.98  & 29.41 \\ 
\hline
\multirow{2}{*}{3-class} 
& \projectName  & \cellcolor{gray!30}33.20  & \cellcolor{gray!30}31.13  & \cellcolor{gray!30}35.43  & \cellcolor{gray!30}37.84  & \cellcolor{gray!30}25.77  & \cellcolor{gray!30}31.90  & \cellcolor{gray!30}32.32  & \cellcolor{gray!30}26.49  & \cellcolor{gray!30}31.76 \\
& MwT  & 43.39 & 38.65 & 50.51 & 68.18  & 30.77  & 43.10  & 39.28  & 42.33  & 44.53 \\ 
\hline
\multirow{2}{*}{4-class} 
& \projectName  & \cellcolor{gray!30}40.07  & \cellcolor{gray!30}38.37  & \cellcolor{gray!30}43.50  & \cellcolor{gray!30}47.30  & \cellcolor{gray!30}30.46  & \cellcolor{gray!30}40.87  & \cellcolor{gray!30}39.32  & \cellcolor{gray!30}34.19  & \cellcolor{gray!30}39.76 \\
& MwT  & 52.93 & 47.00 & 65.12 & 70.38  & 33.80  & 51.49  & 53.46  & 45.64  & 52.48 \\ 
\hline
\multirow{2}{*}{5-class} 
& \projectName  & \cellcolor{gray!30}47.31  & \cellcolor{gray!30}46.61  & \cellcolor{gray!30}50.25  & \cellcolor{gray!30}54.01  & \cellcolor{gray!30}34.78  & \cellcolor{gray!30}48.15  & \cellcolor{gray!30}44.67  & \cellcolor{gray!30}40.18  & \cellcolor{gray!30}45.99 \\
& MwT  & 58.29 & 53.15 & 71.41 & 72.28  & 37.67  & 56.97  & 56.30  & 47.18  & 56.65 \\ 
\hline
\multirow{2}{*}{6-class} 
& \projectName  & \cellcolor{gray!30}52.38  & \cellcolor{gray!30}53.39  & \cellcolor{gray!30}54.90  & \cellcolor{gray!30}60.45  & \cellcolor{gray!30}37.72  & \cellcolor{gray!30}54.50  & \cellcolor{gray!30}49.67  & \cellcolor{gray!30}44.96  & \cellcolor{gray!30}50.87 \\
 & MwT  & 63.75 & 57.31 & 74.58 & 74.01  & 41.25  & 64.69  & 58.62  & 49.34  & 60.94 \\ 
 \hline
\multirow{2}{*}{7-class} 
& \projectName  & \cellcolor{gray!30}59.81  & \cellcolor{gray!30}58.46  & \cellcolor{gray!30}60.68  & \cellcolor{gray!30}65.33  & \cellcolor{gray!30}42.62  & \cellcolor{gray!30}60.10  & \cellcolor{gray!30}55.94  & \cellcolor{gray!30}48.11  & \cellcolor{gray!30}56.88 \\
& MwT  & 72.57 & 61.27 & 79.79 & 78.36  & 52.00  & 68.75  & 66.45  & 51.04  & 66.28 \\ 
\hline
\multirow{2}{*}{8-class} 
& \projectName  & \cellcolor{gray!30}66.90  & 65.29  & \cellcolor{gray!30}66.39  & \cellcolor{gray!30}70.25  & \cellcolor{gray!30}47.02  & \cellcolor{gray!30}67.50  & \cellcolor{gray!30}60.39  & \cellcolor{gray!30}50.85  & \cellcolor{gray!30}61.95 \\
 & MwT  & 79.88 & \cellcolor{gray!30}63.65 & 82.35 & 79.56  & 57.14  & 72.55  & 68.37  & 52.37  & 69.73 \\ 
 \hline
\multirow{2}{*}{9-class} 
& \projectName  & \cellcolor{gray!30}74.86  & 69.93  & \cellcolor{gray!30}72.89  & \cellcolor{gray!30}74.58  & \cellcolor{gray!30}52.57  & \cellcolor{gray!30}73.39  & \cellcolor{gray!30}68.92  & \cellcolor{gray!30}55.04  & \cellcolor{gray!30}67.40 \\
 & MwT  & 88.34 & \cellcolor{gray!30}66.25 & 86.87 & 80.60  & 67.52  & 75.94  & 72.44  & 53.72  & 74.46 \\ 
 \hline
\multirow{2}{*}{10-class}
& \projectName  & \cellcolor{gray!30}82.72  & 74.40  & \cellcolor{gray!30}77.77  & \cellcolor{gray!30}77.91  & \cellcolor{gray!30}57.16  & \cellcolor{gray!30}78.10  & \cellcolor{gray!30}73.67  & \cellcolor{gray!30}57.50  & \cellcolor{gray!30}72.65 \\
& MwT  & 94.11 & \cellcolor{gray!30}68.61 & 88.75 & 81.52  & 72.40  & 78.82  & 74.22  & 54.71  & 76.77 \\ 
\bottomrule
\end{tabular}
}
\end{table}
\begin{table}[t]
\caption{The test accuracy results of \projectName in reusing CNN modules. All results in \%.}
\label{tab:rq3_mwt_and_ours_cnn_acc}
\vspace{-6pt}
\centering
\resizebox{0.8\columnwidth}{!}{
\begin{tabular}{ccrrrrrrrrr}
\toprule
\multirow{2}{*}{\textbf{\begin{tabular}[c]{@{}c@{}}Target\\Task\end{tabular}}} & \multirow{2}{*}{\textbf{Approach}}  & \multicolumn{2}{c}{\textbf{VGG16}} & \multicolumn{2}{c}{\textbf{ResNet18}} & \multicolumn{2}{c}{\textbf{SimCNN}} & \multicolumn{2}{c}{\textbf{ResCNN}} & \multirow{2}{*}{\textbf{Average}}
\\ \cmidrule(lr){3-4} \cmidrule(lr){5-6} \cmidrule(lr){7-8} \cmidrule(lr){9-10} &  
& \textbf{CIFAR10}  & \textbf{SVHN}  & \textbf{CIFAR10}  & \textbf{SVHN}  & \textbf{CIFAR10}  & \textbf{SVHN}  & \textbf{CIFAR10}    & \textbf{SVHN}   &  
\\ \midrule \midrule
\multirow{3}{*}{2-class}  
& MwT  & 99.20  & 99.17  & 99.30  & 99.33  & 99.15  & 99.06  & 99.10  & 99.09  & 99.18  \\  
& \projectName  & 99.15  & 98.85  & 99.40  & 99.21  & 98.95  & 99.01  & 99.10  & 98.93  & 99.08  
\\ \cline{2-11}
& \textbf{loss}  & 0.05  & 0.32  & \cellcolor{gray!30}-0.10  & 0.12  & 0.20  & 0.05  & 0.00  & 0.16  & 0.10  
\\ \hline
\multirow{3}{*}{3-class}  
& MwT  & 97.40  & 98.58  & 97.23  & 98.82  & 96.83  & 98.63  & 97.77  & 98.69  & 97.99  \\
& \projectName  & 97.50  & 98.56  & 97.10  & 98.48  & 97.07  & 98.46  & 97.60  & 98.34  & 97.89
\\ \cline{2-11}
& \textbf{loss}  & \cellcolor{gray!30}-0.10  & 0.02  & 0.13  & 0.34  & \cellcolor{gray!30}-0.24  & 0.17  & 0.17  & 0.35  & 0.10  
\\ \hline
\multirow{3}{*}{4-class}  
& MwT  & 94.98  & 97.28  & 95.50  & 97.64  & 94.65  & 96.90  & 95.33  & 97.13  & 96.18  \\
& \projectName  & 95.40  & 97.74  & 95.00  & 97.06  & 94.90  & 96.40  & 95.43  & 96.28  & 96.03 
\\ \cline{2-11}
& \textbf{loss}  & \cellcolor{gray!30}-0.42  & \cellcolor{gray!30}-0.46  & 0.50  & 0.58  & \cellcolor{gray!30}-0.25  & 0.50  & \cellcolor{gray!30}-0.10  & 0.85  & 0.15 
\\ \hline
\multirow{3}{*}{5-class}  
& MwT  & 93.82  & 97.01  & 94.34  & 97.55  & 92.50  & 96.99  & 93.96  & 96.54  & 95.34  \\
& \projectName  & 93.92  & 97.15  & 93.58  & 97.19  & 93.58  & 96.83  & 94.18  & 96.34  & 95.35
\\ \cline{2-11}
& \textbf{loss}  & \cellcolor{gray!30}-0.10  & \cellcolor{gray!30}-0.14  & 0.76  & 0.36  & \cellcolor{gray!30}-1.08  & 0.16  & \cellcolor{gray!30}-0.22  & 0.20  & \cellcolor{gray!30}-0.01
\\ \hline
\multirow{3}{*}{6-class}  
& MwT  & 91.05  & 96.44  & 91.27  & 96.92  & 89.80  & 96.34  & 90.90  & 95.84  & 93.57  \\
& \projectName  & 90.72  & 96.28  & 91.13  & 96.58  & 90.68  & 96.25  & 90.92  & 95.80  & 93.54
\\ \cline{2-11}
& \textbf{loss}  & 0.33  & 0.16  & 0.14  & 0.34  & \cellcolor{gray!30}-0.88  & 0.09  & \cellcolor{gray!30}-0.02  & 0.04  & 0.03 
\\ \hline
\multirow{3}{*}{7-class}  
& MwT  & 90.90  & 95.78  & 91.86  & 96.45  & 89.74  & 95.90  & 90.56  & 95.66  & 93.36 \\
& \projectName  & 90.84  & 96.23  & 90.83  & 96.42  & 90.40  & 96.22  & 90.83  & 95.04  & 93.48 
\\ \cline{2-11}
& \textbf{loss}  & 0.06  & \cellcolor{gray!30}-0.45  & 1.03  & 0.03  & \cellcolor{gray!30}-0.66  & \cellcolor{gray!30}-0.32  & \cellcolor{gray!30}-0.27  & 0.62  & \cellcolor{gray!30}-0.12
\\ \hline
\multirow{3}{*}{8-class}  
& MwT  & 91.03  & 95.52  & 91.71  & 96.25  & 89.14  & 95.26  & 90.59  & 95.16  & 93.08 \\
& \projectName  & 90.76  & 95.64  & 90.19  & 95.99  & 90.39  & 95.50  & 90.91  & 95.09  & 93.06
\\ \cline{2-11}
& \textbf{loss} & 0.27  & \cellcolor{gray!30}-0.12  & 1.52  & 0.26  & \cellcolor{gray!30}-1.25  & \cellcolor{gray!30}-0.24  & \cellcolor{gray!30}-0.32  & 0.07  & 0.02
\\ \hline
\multirow{3}{*}{9-class}  
& MwT  & 90.89  & 95.20  & 91.48  & 96.00  & 89.34  & 95.02  & 90.59  & 94.88  & 92.93  \\
& \projectName  & 90.84  & 95.75  & 90.11  & 95.81  & 90.24  & 95.28  & 90.73  & 94.41  & 92.90  
\\ \cline{2-11}
& \textbf{loss}  & 0.05  & \cellcolor{gray!30}-0.55  & 1.37  & 0.19  & \cellcolor{gray!30}-0.90  & \cellcolor{gray!30}-0.26  & \cellcolor{gray!30}-0.14  & 0.47  & 0.03
\\ \hline
\multirow{3}{*}{10-class}  
& MwT  & 90.86  & 94.74  & 91.59  & 95.95  & 89.54  & 94.95  & 90.35  & 94.36  & 92.79  \\
& \projectName  & 90.69  & 95.46  & 90.36  & 95.52  & 90.50  & 95.04  & 90.41  & 94.37  & 92.79
\\ \cline{2-11}
& \textbf{loss}  & 0.17  & \cellcolor{gray!30}-0.72  & 1.23  & 0.43  & \cellcolor{gray!30}-0.96  & \cellcolor{gray!30}-0.09  & \cellcolor{gray!30}-0.06  & \cellcolor{gray!30}-0.01  & \cellcolor{gray!30}0.00
\\ \bottomrule
\end{tabular}
}
\end{table}
\begin{table}[t]
\setlength\tabcolsep{3pt}
\newcommand{\reductionpercentages}[2]{%
  \raisebox{-0.5ex}{\scriptsize\textcolor{blue}{$\downarrow$#1\%},\textcolor{red}{$\downarrow$#2\%}}%
}
\caption{The comparison of \projectName and MwT for FLOPs in reusing CNN modules, all results in FLOPs(M).}
\label{tab:rq3_flops_cnns}
\vspace{-6pt}
\centering
\resizebox{0.8\columnwidth}{!}{
\begin{tabular}{c c c c c c c c c c c}
\toprule
\multirow{2}{*}{\textbf{Target Task}} & \multirow{2}{*}{\textbf{Approach}} 
& \multicolumn{2}{c}{\textbf{VGG16(314.28)}}  & \multicolumn{2}{c}{\textbf{ResNet18(558.59)}}  & \multicolumn{2}{c}{\textbf{SimCNN(313.73)}}  & \multicolumn{2}{c}{\textbf{ResCNN(431.17)}}  & \multirow{2}{*}{\textbf{Average(404.44)}} 
\\ 
\cmidrule(lr){3-4} \cmidrule(lr){5-6} \cmidrule(lr){7-8} \cmidrule(lr){9-10} &  
& \textbf{CIFAR10}   & \textbf{SVHN}  & \textbf{CIFAR10}   & \textbf{SVHN}  & \textbf{CIFAR10}   & \textbf{SVHN}  & \textbf{CIFAR10}   & \textbf{SVHN}  &
\\ \midrule \midrule
\multirow{2}{*}{2-class} 
& \projectName   & 87.11  & 35.77  & 159.35  & 75.25  & 75.86  & 48.01  & 100.60  & 55.64  & 79.70\reductionpercentages{0.85}{80.29} \\
& MwT  & 70.97 & 31.85 & 129.76 & 136.15  & 48.95  & 82.24  & 82.52  & 60.59  & 80.38 \\ 
\hline
\multirow{2}{*}{3-class} 
& \projectName  & 120.94  & 36.12 & 261.50  & 105.43  & 98.72  & 61.02  & 122.90  & 59.08  & 108.21\reductionpercentages{21.11}{73.24} \\
& MwT  & 121.21 & 52.93 & 189.50 & 329.49  & 70.71  & 98.99  & 130.55  & 103.95  & 137.17 \\ 
\hline
\multirow{2}{*}{4-class} 
& \projectName  & 143.95  & 36.14  & 325.78  & 142.29  & 121.93  & 83.59  & 146.72  & 65.71  & 133.26\reductionpercentages{25.46}{67.05} \\
& MwT  & 166.35 & 73.31 & 336.49 & 336.41  & 90.85  & 138.09  & 184.12  & 104.54  & 178.77 \\ 
\hline
\multirow{2}{*}{5-class} 
& \projectName  & 165.66  & 36.59  & 371.72  & 213.50  & 133.49  & 98.43  & 156.78  & 69.80  & 155.75\reductionpercentages{17.88}{61.49} \\
& MwT  & 179.99 & 83.82 & 360.62 & 347.17  & 97.23  & 146.82  & 196.81  & 104.87  & 189.67 \\ 
\hline
\multirow{2}{*}{6-class} 
& \projectName  & 179.72  & 36.63  & 404.64  & 235.81  & 142.30  & 112.24  & 170.12  & 74.91  & 169.55\reductionpercentages{17.50}{58.08} \\
 & MwT  & 204.39 & 91.03 & 376.75 & 382.28  & 111.61  & 175.23  & 197.70  & 105.17  & 205.52 \\ 
 \hline
\multirow{2}{*}{7-class} 
& \projectName  & 202.36  & 36.68  & 435.54  & 259.49  & 158.24  & 124.24  & 191.31  & 77.74  & 185.70\reductionpercentages{17.90}{54.08} \\
& MwT  & 230.41 & 99.47 & 398.82 & 372.42  & 154.26  & 203.62  & 235.47  & 114.96  & 226.18 \\ 
\hline
\multirow{2}{*}{8-class} 
& \projectName  & 222.99  & 36.95  & 466.13  & 283.41  & 170.16  & 143.83  & 198.28  & 79.17  & 200.12\reductionpercentages{15.15}{50.52} \\
 & MwT  & 249.33 & 103.30 & 407.57 & 377.54  & 174.49  & 221.60  & 237.63  & 115.22  & 235.84 \\ 
 \hline
\multirow{2}{*}{9-class} 
& \projectName  & 246.74  & 36.99  & 502.10  & 303.25  & 184.69  & 161.55  & 216.78  & 83.81  & 217.49\reductionpercentages{15.18}{46.22} \\
 & MwT  & 279.40 & 107.36 & 432.79 & 382.75  & 219.00  & 245.28  & 268.31  & 116.28  & 256.40 \\ 
 \hline
\multirow{2}{*}{10-class}
& \projectName  & 272.56  & 37.03  & 529.28  & 318.6  & 197.23  & 175.50  & 225.51  & 86.39  & 230.26\reductionpercentages{12.52}{43.07} \\
& MwT  & 293.88 & 111.10 & 442.67 & 388.09  & 231.84  & 252.71  & 268.87  & 116.62  & 263.22 \\ 
\bottomrule
\end{tabular}
}
\end{table}

\textbf{On-demand Reuse.}
To compare \projectName with MwT in on-demand model reuse, we randomly select 10 sub-tasks for each n-class classification scenario.
In each n-class classification scenario, we reuse the module on demand from MwT and \projectName, then analyze the average number of convolution kernels, FLOPs(M) and the accuracy of the models.

Table ~\ref{tab:rq2_mwt_and_ours_krr} compares \projectName with MwT in reusing CNN models in terms of kernel retention rate (KRR). On average, \projectName significantly reduces KRR in the modules. For each sub-task, \projectName is capable of generating smaller modules compared to MwT. We further compared the accuracy of on-demand reuse in Table ~\ref{tab:rq3_mwt_and_ours_cnn_acc}.
On average, \projectName maintains performance parity with MwT, exhibiting minimal to no degradation in effectiveness. Notably, the modules generated by \projectName are significantly smaller than those produced by MwT. Given this substantial reduction in module size and competitive performance, we can conclude that \projectName demonstrates superior effectiveness in the on-demand reuse of CNN models compared to MwT. This improvement in efficiency without compromising performance underscores \projectName's potential to advance modular approaches in deep learning applications.

Table~\ref{tab:rq3_flops_cnns} evaluates the computational overhead reduction achieved by \projectName and MwT. The numbers in the model name row represent the FLOPs(M) count when directly reusing standard trained models. Our experiments show that \projectName significantly simplifies the on-demand reuse of CNN models, reducing FLOPs by up to 80.29\%. Compared to MwT, \projectName also demonstrates improvement in FLOPs reduction, decreasing computational costs by up to 25.46\%.

\begin{tcolorbox}[colback=white, colframe=black, boxrule=0.4mm, arc=2mm, left=3mm, right=3mm, top=1mm, bottom=1mm, width=\columnwidth]
\projectName is generalizable to CNNs.
Compared with MwT on four CNN models, \projectName not only achieves 0.18\%, 0.0198, 0.0352, and 4.75\% improvements on average accuracy, cohesion, coupling, and KRR in modular training but also achieves higher accuracy, lower module size, and up to 25.46\% fewer FLOPs in on-demand reuse.
\end{tcolorbox}

\subsubsection{RQ4 - Impact of hyper-parameters}

\begin{figure}[t]
    \centering
    \begin{subfigure}[b]{0.24\textwidth}
        \centering
        \includegraphics[width=\textwidth]{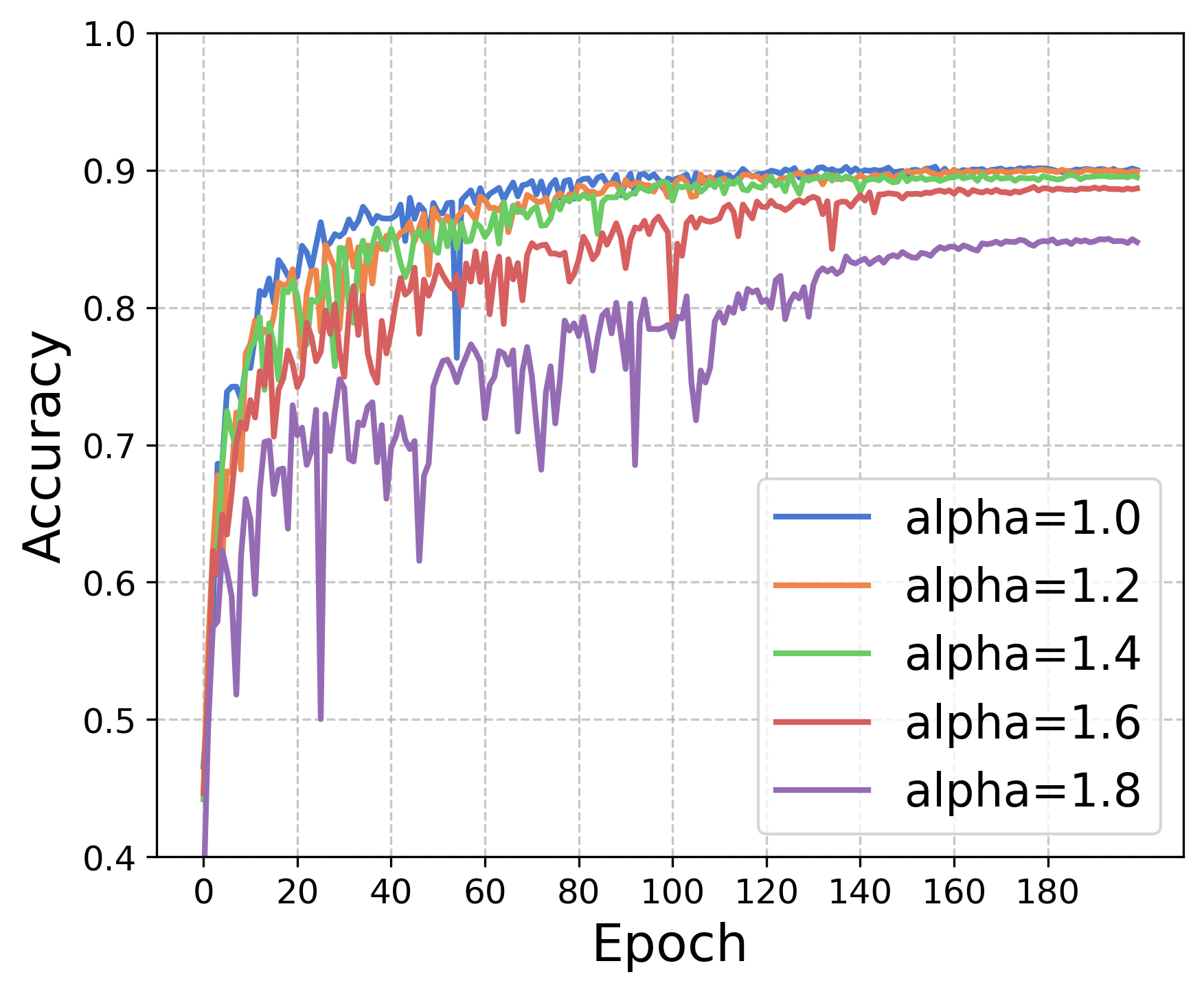}
    \end{subfigure}
    \begin{subfigure}[b]{0.24\textwidth}
        \centering
        \includegraphics[width=\textwidth]{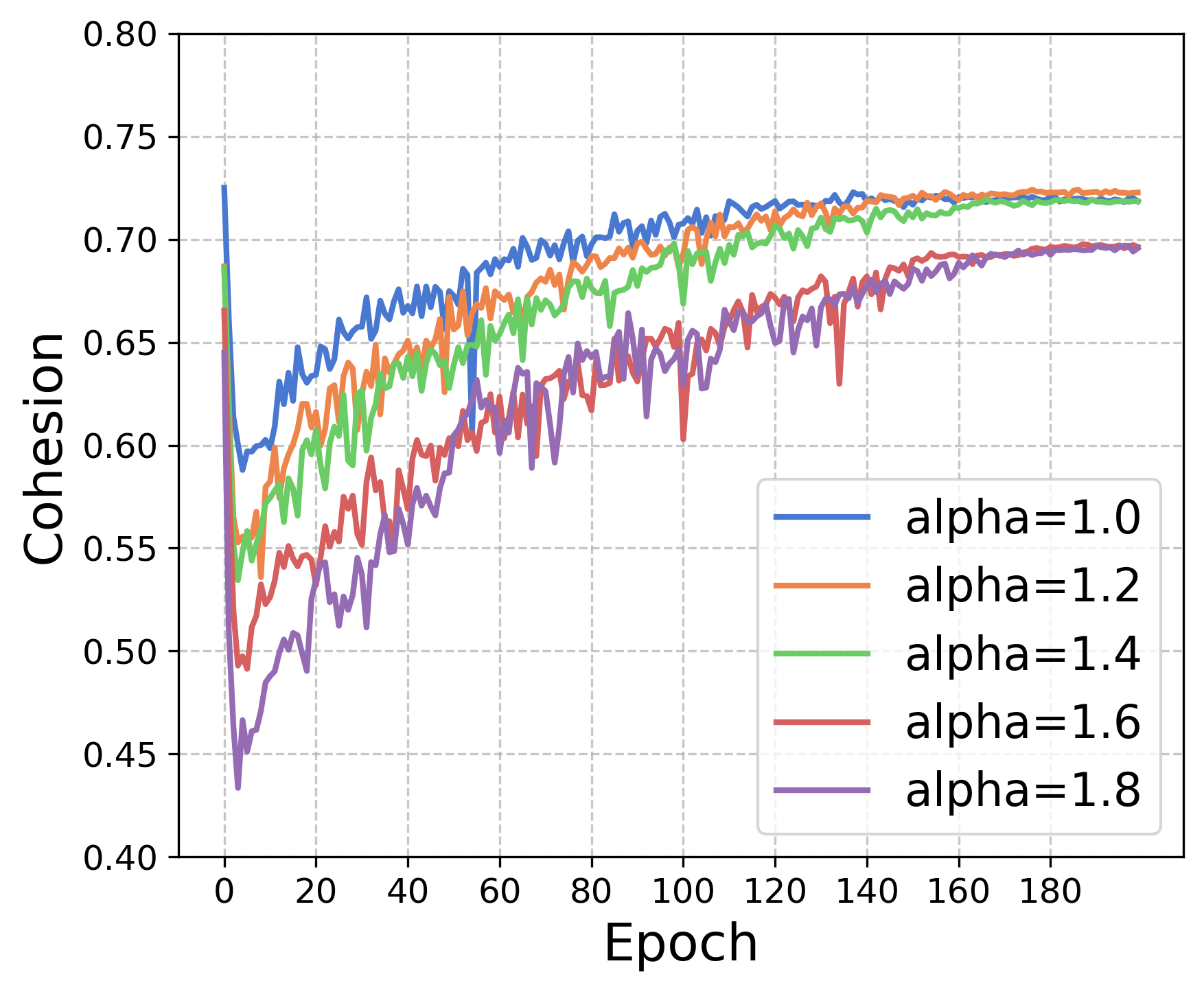}
    \end{subfigure}
    \begin{subfigure}[b]{0.24\textwidth}
        \centering
        \includegraphics[width=\textwidth]{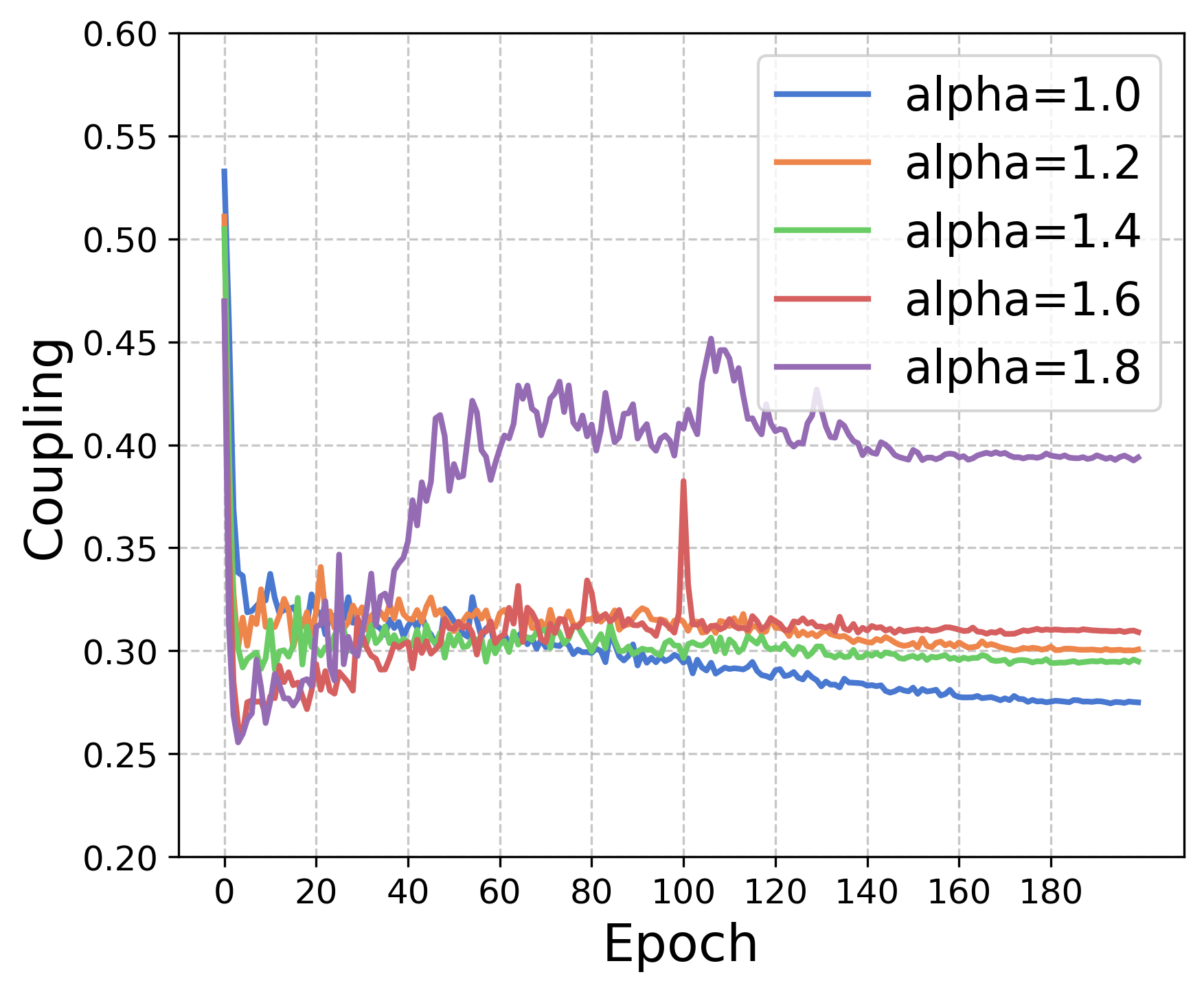}
    \end{subfigure}
    \begin{subfigure}[b]{0.24\textwidth}
        \centering
        \includegraphics[width=\textwidth]{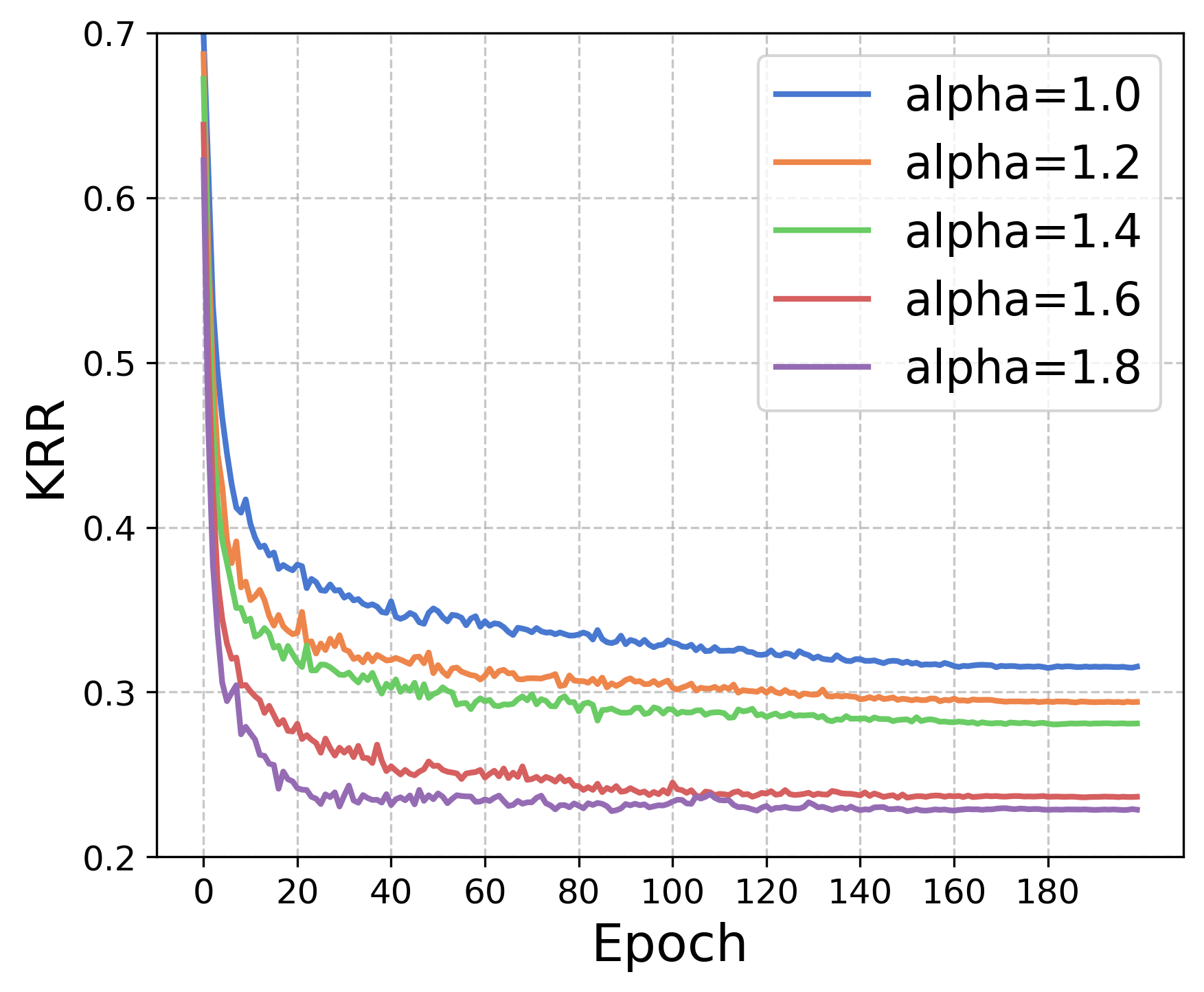}
    \end{subfigure}
    \caption{The modular training process when $\alpha$ ranges from 1.0 to 1.8.}
    \label{fig:hyperparameters_alpha}
\end{figure}

\begin{figure}[t]
    \centering
    \begin{subfigure}[b]{0.24\textwidth}
        \centering
        \includegraphics[width=\textwidth]{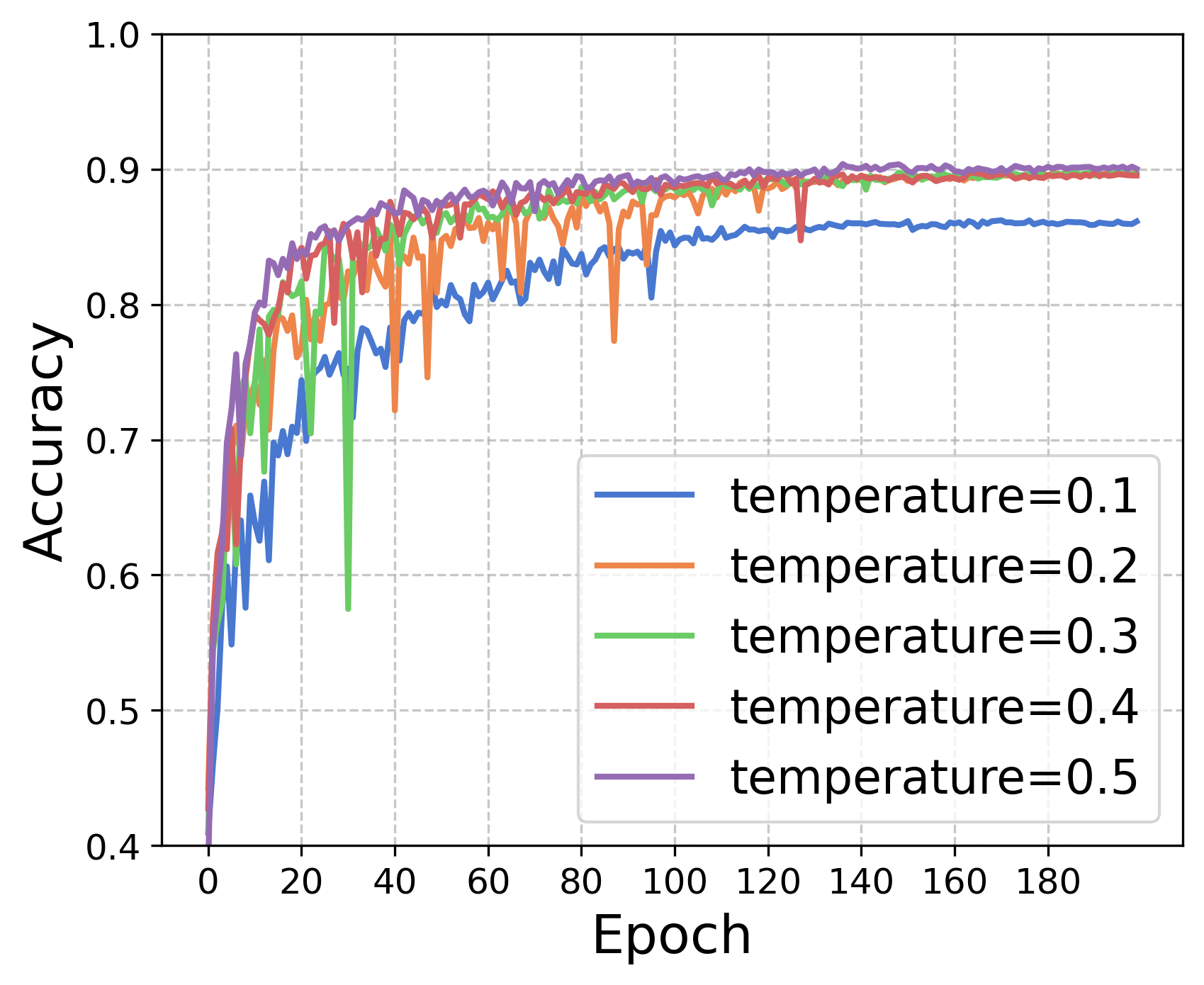}
    \end{subfigure}
    \begin{subfigure}[b]{0.24\textwidth}
        \centering
        \includegraphics[width=\textwidth]{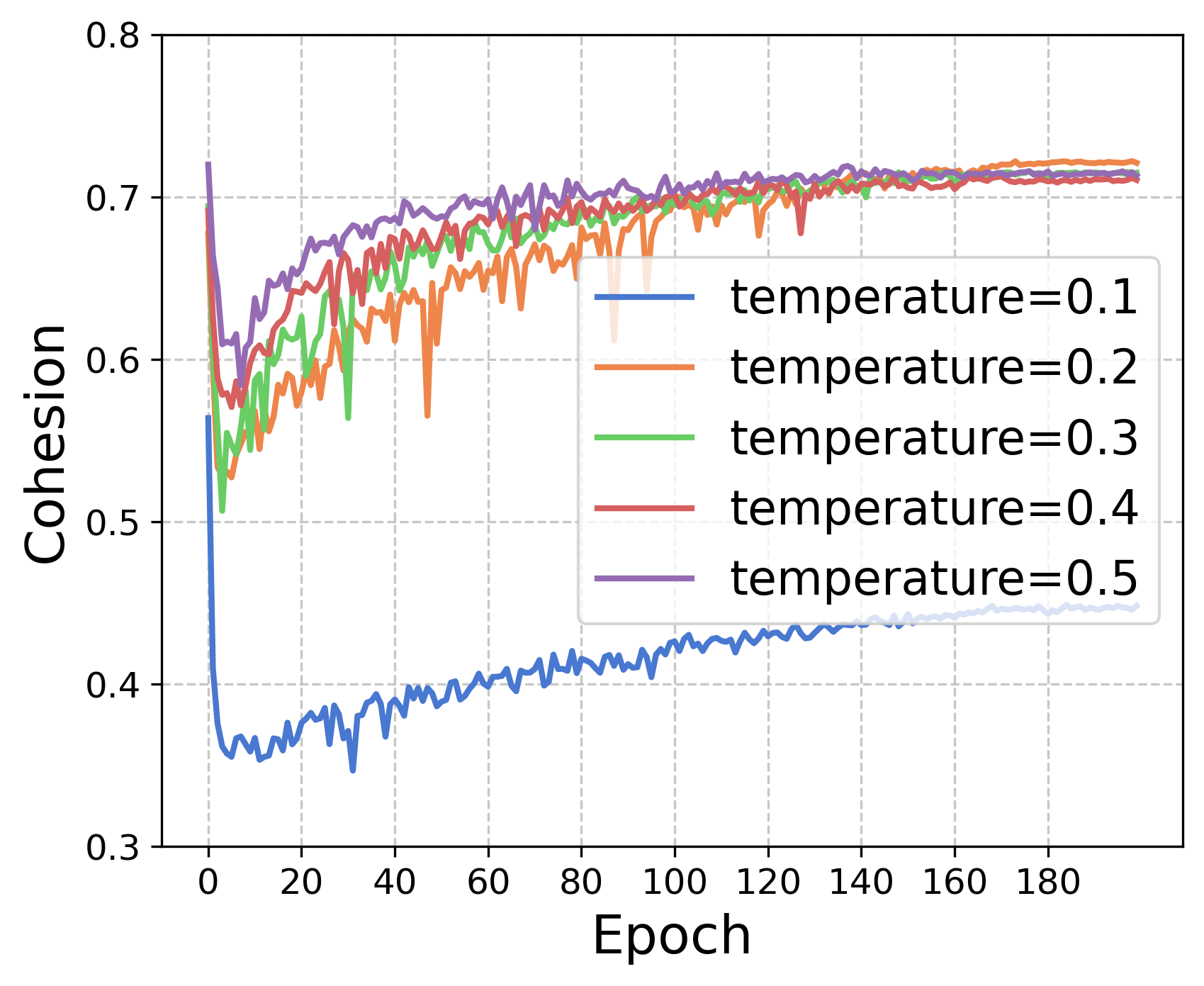}
    \end{subfigure}
    \begin{subfigure}[b]{0.24\textwidth}
        \centering
        \includegraphics[width=\textwidth]{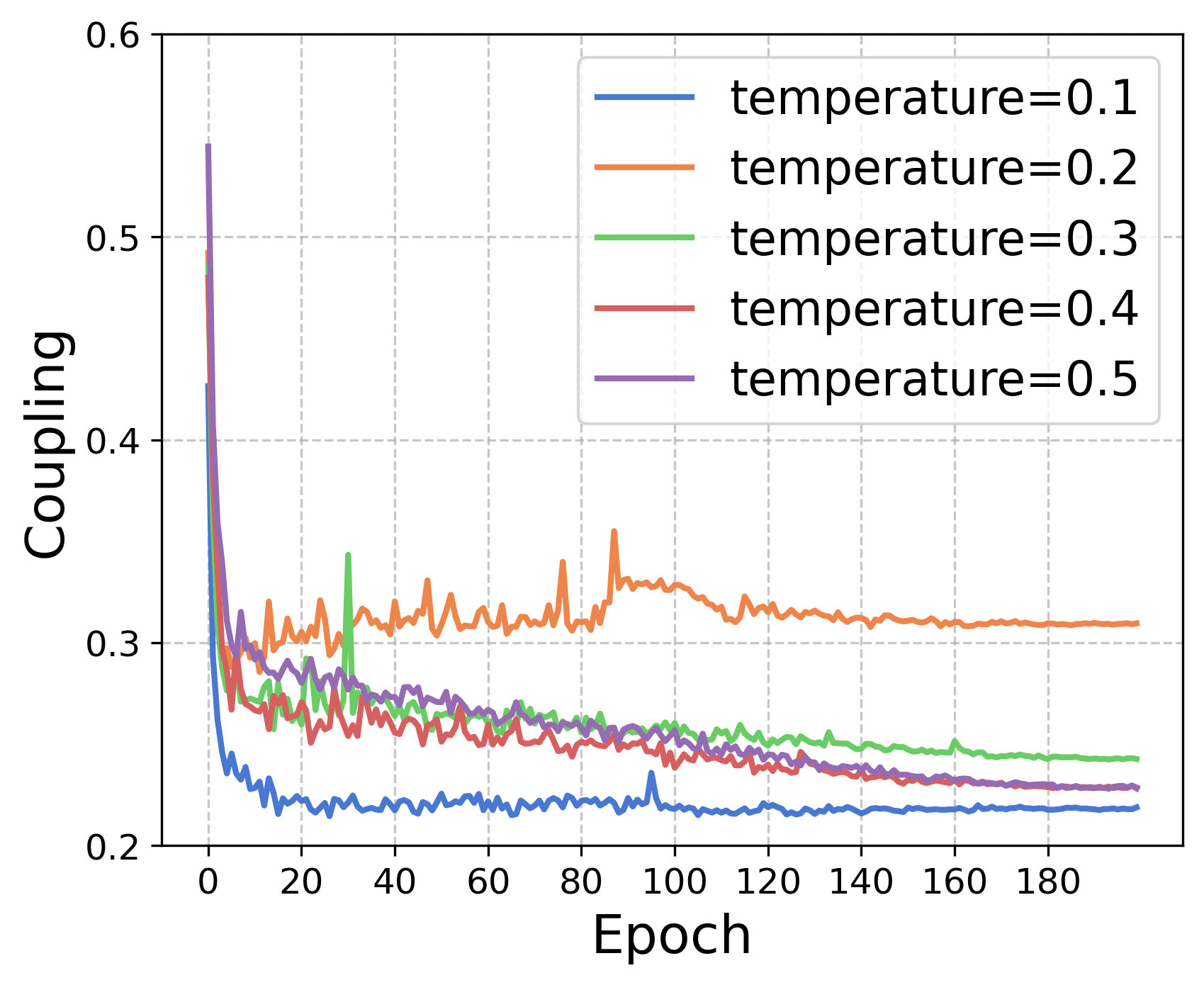}
    \end{subfigure}
    \begin{subfigure}[b]{0.24\textwidth}
        \centering
        \includegraphics[width=\textwidth]{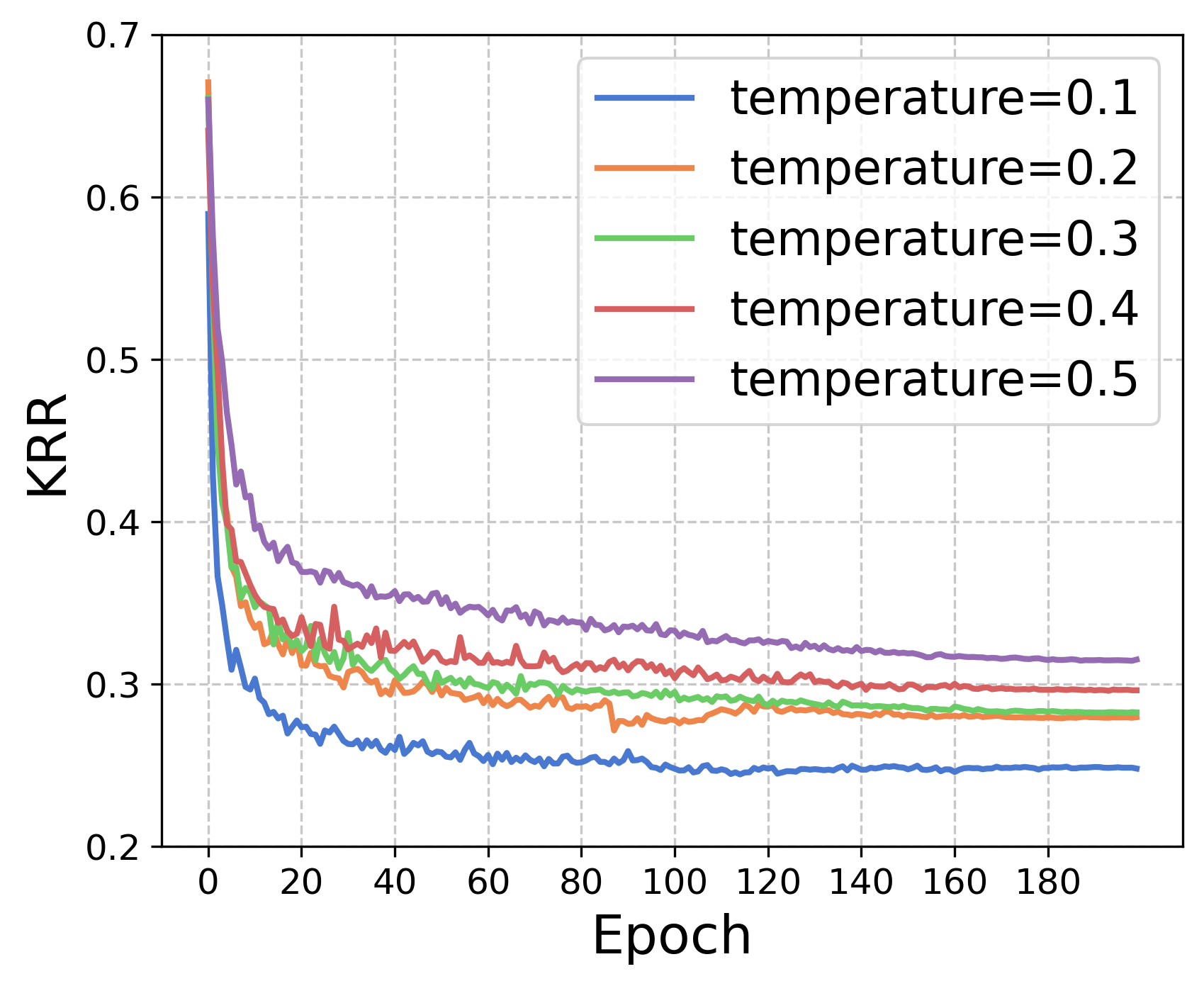}
    \end{subfigure}
    \caption{The modular training process when $\tau$ ranges from 0.1 to 0.5.}
    \label{fig:hyperparameters_tau}
\end{figure}

Finally, we investigate the influence of hyper-parameter $\alpha$ (the weighting factor of contrastive loss) during the training process. Additionally, we also discussed $\tau$, which is the temperature in the contrastive loss. Due to the diversity in model architectures, we only demonstrated the training process of the VGG16 model on CIFAR-10. More experimental results and training details can be found on the project webpage ~\cite{bi2024nemo}.

We observed that $\alpha$ directly influences the modular training process. Figure ~\ref{fig:hyperparameters_alpha} illustrates the accuracy, coupling degree, cohesion degree, and KRR during the modular training process, where $\alpha$ values are set to 1.0, 1.2, 1.4, 1.6, and 1.8, with a consistent batch size of 128. After 200 epochs of training, it is evident that as $\alpha$ increases, the model accuracy gradually decreases, but the KRR also significantly declines. Within a reasonable range, moderately increasing $\alpha$ does not lead to substantial accuracy loss but can significantly reduce the KRR. However, excessively large $\alpha$  values make the model difficult to fit and result in unacceptable accuracy loss.

As a crucial hyperparameter in contrastive learning, the temperature 
$\tau$ significantly affects the effectiveness of modular training. Theoretically, a smaller $\tau$ sharpens the sample distribution, making the training process more focused on distinguishing features of specific hard samples. Conversely, a larger $\tau$ smooths the sample distribution, making the training more focused on overall features. As shown in Fig.~\ref{fig:hyperparameters_tau}, when $\tau$ = 0.1, which is a very low value, the model's cohesion becomes very low. This means that the neuron recognizers focus more on the features of specific samples rather than using similar convolutional kernels for similar samples. On the other hand, when $\tau$ = 0.5, the sample distribution becomes smoother, and the neuron recognizers focus more on overall features, resulting in the selection of more neurons to ensure that all similar samples use similar neurons.
It is noteworthy that $\tau$ not only affects the sample distribution but also changes the absolute value of the whole contrastive loss, making it difficult to compare different $\tau$ values with the same $\alpha$. However, based on our experiments, $\tau$ is not a sensitive parameter for modular training process, and for most models and datasets, a $\tau$ value between 0.2 and 0.3 yields good results. Therefore, we chose to use $\tau$ = 0.2.

\begin{tcolorbox}[colback=white, colframe=black, boxrule=0.4mm, arc=2mm, left=3mm, right=3mm, top=1mm, bottom=1mm, width=\columnwidth]
The impact of \projectName's hyperparameters on performance is predictable. Within a reasonable range, $\tau$ requires no adjustment. $\alpha$ is negatively correlated with kernel or neuron retention rate and, within certain limits, has almost no effect on accuracy, making \projectName easy to apply to new models.
\end{tcolorbox}
\subsubsection{RQ5 - Scalability of \projectName}
Compared to previous work MwT, \projectName provides neuron-level identification and modularization methods. However, for most models, the number of neurons is typically much larger than the number of convolution kernels in CNN models with comparable parameter counts. For instance, in our experiments, CNN models had an average of 4,160 convolution kernels, while the ViT-small model contained over 40,000 neurons. This quantity directly affects the number of our optimization targets. Since neurons are present in all neural network models, \projectName will be extended to more model architectures in future work, which requires our loss function to maintain effectiveness when facing larger optimization targets. We redesigned the modularization training loss function based on contrastive learning and metrics cohesion and coupling. To evaluate the effectiveness of this new loss function, we selected three ViT models with different neuron counts and evaluated performance by only varying the loss function.

As shown in Table~\ref{tab:rq5-scalibility}, with increasing neuron counts in ViT models, MwT struggles significantly more than \projectName to balance neuron retention rate (NRR) and accuracy (ACC) loss. MwT either suffers excessive accuracy degradation (-1.39) or maintains several times more neurons (+25.33) to preserve acceptable accuracy. Given that ACC and NRR represent a clear tradeoff, our experiments demonstrate that MwT underperforms in scenarios with more optimization objectives and typically results in worse accuracy. 
\begin{table}[t]
\caption{Scalibility of \projectName.}
\label{tab:rq5-scalibility}
\vspace{-6pt}
\centering
\resizebox{0.7\columnwidth}{!}{
\begin{tabular}{c c c c c c c c c c}
\toprule
\multirow{2}{*}{\textbf{Model}} & \multirow{2}{*}{\textbf{Approach}} & \multirow{2}{*}{\textbf{\#Neurons}} & \multirow{2}{*}{\textbf{\begin{tabular}[c]{@{}c@{}}Modular\\ Model ACC\end{tabular}}} & \multicolumn{6}{c}{\textbf{Modules}}       
\\ \cmidrule(lr){5-10} &    &    &    & \textbf{NRR} & \textbf{Cohesion} & \textbf{Coupling} & \textbf{$\alpha$} & \textbf{$\beta$} & \textbf{$\tau$} 
\\ \midrule \midrule
\multirow{2}{*}{$ViT_{tiny}$} 
& \projectName  & \multirow{2}{*}{13824}  & 90.37  & 15.65  & 0.9633  & 0.1941  & 0.2  & -  & 0.3   
\\  & MwT  &  & 89.66 \textbf{(-0.71)}  & 33.71 \textbf{(+18.06)}  & 0.9736  & 0.3076  & 0.5  & 0.5  & -  
\\ \midrule
\multirow{2}{*}{$ViT_{small}$} 
& \projectName  & \multirow{2}{*}{41472}  & 91.58  & 12.04  & 0.9283  & 0.0899  & 0.2  & -  & 0.3   
\\  & MwT  &  & 90.19 \textbf{(-1.39)}  & 20.81 \textbf{(+8.77)}  & 0.9414  & 0.1866  & 0.1  & 0.5  & -  
\\ \midrule
\multirow{2}{*}{$ViT_{base}$} 
& \projectName  & \multirow{2}{*}{82994}  & 91.23  & 15.47  & 0.9612  & 0.2105  & 0.2  & -  & 0.3   
\\  & MwT  &  & 90.83 \textbf{(-0.40)}  & 40.80 \textbf{(+25.33)}  & 0.9902  & 0.4414  & 0.5  & 0.5  & -  
\\ \bottomrule
\end{tabular}
}
\end{table}

\subsection{Case Study}
\label{sec:case_study}
One potential application of \projectName is to enable model sharing platforms to provide on-demand model reuse~\cite{modelfoundry}.
Unlike existing model sharing platforms (e.g., HuggingFace) that only support entire model reuse, platforms equipped with on-demand model reuse functionality allow developers to reuse only the relevant modules.
This approach mirrors practices in software engineering where fine-grained modularity can reduce the reuse overhead faced by users.
A recent work, ModelFoundry~\cite{modelfoundry}, has explored this scenario. 
ModelFoundry established a modularization and composition system, integrating several modularization algorithms~\cite{SeaM, GradSplitter} to decompose models into functional modules.
It maintains a repository of these modules with search and cost estimation capabilities for on-demand reuse.
Currently, ModelFoundry supports only reusing CNN models due to modularization algorithms' limitations. 
\projectName can extend the system to support the on-demand reuse of Transformer models further.

\subsubsection{Application to Vision-related Tasks}

Specifically, in typical development scenarios, developers often reuse an entire pretrained model, even when only a subset of its functionality is required.~\cite{spoof_detect,bone_detection,rock_paper_scissors}.
For example, the Rock-Paper-Scissors~\cite{rock_paper_scissors} project reuses a ViT model pretrained on ImageNet and fine-tunes it on a three-class classification dataset including ``rock'', ``paper'', and ``scissors'' images.
Although the downstream task contains only three classes, the project reuses the entire ViT model with a lot of redundant weights corresponding to the irrelevant classes, incurring additional reuse overhead.
\projectName can alleviate the problem by pre-training the model using modular training.
Thus, the project can reuse only the relevant module, such as the module identifying hand-related images, and fine-tune the module on the rock-paper-scissors dataset, resulting in a much smaller model with lower inference overhead.

We experimentally evaluate \projectName in this case, demonstrating \projectName's effectiveness in practical scenarios.
Due to the huge training overhead of the ViT model on the ImageNet dataset (several days or weeks), we construct a small dataset for modular training, including 5000 airplane and 5000 automobile images from CIFAR-10, and 7500 hand images from Sign Language MNIST~\cite{sign_language_mnist}.
We pre-train a ViT-small model using \projectName on our constructed dataset and fine-tune the module corresponding to the ``hand'' class on the Rock-Paper-Scissors dataset~\cite{rock_paper_scissors}.
For a fair comparison, we pre-train a ViT-small model using the standard training method and fine-tune it on the Rock-Paper-Scissors dataset.
Table~\ref{tab:case_study} shows the results in terms of accuracy and NRR.
Compared to reusing the entire model, reusing the module can achieve comparable accuracy while retaining only 47.6\% of neurons. The reduction of 62.54\% FLOPs also indicates a less inference overhead.

\begin{table}[t]
\caption{Performance of \projectName on the Rock Paper Scissors case.}
\label{tab:case_study}
\vspace{-6pt}
\centering
\resizebox{0.6\columnwidth}{!}{
\begin{tabular}{c c c c c c}
\toprule
\textbf{Model} & \textbf{Dataset} & \textbf{Accuracy} & \textbf{NRR} & \textbf{Type} & \textbf{FLOPs (M)} \\
\midrule
ViT-small & Rock Paper Scissors & 72.31\% & 1.000 & Baseline & 1384.49 \\
\midrule
ViT-small module & Rock Paper Scissors & 72.85\% & 0.476 & \projectName & 518.65\rlap{\,\scriptsize\textcolor{red}{$\downarrow$62.54\%}} \\
\bottomrule
\end{tabular}
}
\end{table}  

\subsubsection{Application to Text-related Tasks}

\label{sec:application_scenario}
We also evaluate \projectName on text-related tasks, especially those related to software engineering.
For example, consider a typical scenario of reusing a pre-trained CodeBERT~\cite{codebert} model for code clone detection. CodeBERT~\cite{codebert} is pre-trained on the CodeSearchNet~\cite{codesearchnet} dataset covering six programming languages (\textit{Go, Java, JavaScript, PHP, Python, Ruby}). 
Suppose a user’s task involves detecting code clones exclusively in \textit{Java} programs. Traditionally, the user must reuse the entire CodeBERT model, even though only its \textit{Java} knowledge is relevant.
In contrast, with \projectName, the developer can pre-train CodeBERT in a modular fashion and share individual modules, each specialized for a specific programming language. 
The user can then reuse only the \textit{Java} module for the clone detection task. 
This module is significantly smaller than the entire model, leading to substantial reductions in inference cost.

We experimentally evaluate the practicality of \projectName in this scenario. Specifically, based on the pre-trained CodeBERT model, we use \projectName to perform modular training on the CodeSearchNet dataset for three epochs, resulting in six language-specific modules with an average neuron retention rate of 39.23\%.
We then reuse the \textit{Java} module and fine-tune it on BigCloneBench~\cite{bigclonebench}, a clone detection dataset consisting exclusively of \textit{Java} code. 
The fine-tuned module achieves 96.55\% accuracy on the test set. Compared to reusing the entire pre-trained CodeBERT model, which achieves 98.57\% accuracy, reusing the \textit{Java} module achieves comparable performance while retaining only 43.58\% of the neurons. This demonstrates the effectiveness of \projectName in significantly reducing inference costs without compromising performance.

\section{Discussion}
\subsection{The Generalizability of \projectName}
Given the rapid advancement of deep learning and the emergence of diverse DNN models, the generalizability of DNN modularization approaches is important. 
We argue that \projectName is generalizable to a diverse range of DNN models from two perspectives: (1) Model architecture: \projectName performs modular training at the neuron level, the fundamental component across all DNN model architectures. This enables \projectName applicable to a wide spectrum of neural network models. 
Moreover, \projectName offers a structured decomposition method at the neuron level, removing irrelevant weights from the weight matrix, which supports flexible and on-demand reusability of various DNN models (e.g., CNN, FCNN, RNN, and Transformer-based models). 
(2) Model scale: \projectName introduces a novel modular training loss based on contrastive learning, ensuring its effectiveness and efficiency in larger-scale models. As model complexity increases, the trainable parameters expand from approximately 4,000 convolutional kernels to over 40,000 neurons --- a scale at which existing methods like MwT~\cite{MwT} struggle to maintain the training and modularization performance. By incorporating contrastive learning optimization, \projectName achieves superior performance on Transformer-based models, effectively addressing the scalability challenges in the modularization of large-scale DNN models.

To evaluate the generalizability of \projectName, we applied it to object detection tasks.
The model to be modularized is Detection Transformer model, (DETR)~\cite{DETR}, which has a ResNet50 backbone and a 12-layer transformer encoder-decoder component. 
The multi-object detection task is a practical yet complex task, challenging \projectName due to simultaneous predictions of various targets and bounding boxes. 
To streamline the implementation, we used the remote sensing dataset RSOD (976 images with 6,950 objects)~\cite{RSOD} with four exclusive categories: aircraft, playgrounds, overpasses, and oiltanks. 
In particular, each image contains multiple instances of objects exclusively of a single class.
The decoder component has a classification head and a box head. For each forward propagation, it has a 100 sequence query input for classification and box results. Applying modularization in the decoder component will cause a significant loss of those results. Moreover, 100 queries are too many for our tasks, which have at most 20 objects in each image, causing extra computation intensity in the modular training process. We set it to 20 and selectively modularized the encoder component.
For evaluation metrics, we adopted mAP@50 to quantify detection precision while assessing modular quality through cohesion-coupling analysis of neural components.

The modular DETR model achieved 44.9\% mAP, indicating only 1.7\% performance loss compared to 46.6\% achieved by standard training. 
Furthermore, the high intermodule cohesion (0.9990) indicates that for the same class object, it uses the same group of neurons. The reduced coupling (0.7801) shows that for different class object detection, it uses some different neurons. 
The results demonstrate that \projectName is potentially applicable to object detection tasks.

We also observed that, compared to classification tasks, the modularization performance in terms of coupling drops on object detection tasks.
The primary factors limiting \projectName's modularization efficacy in DETR-based object detection include:
1) The RSOD dataset's limited scale (976 images) and resolution prove insufficient for comprehensive feature learning; 2) Inherent complexity of remote sensing objects challenges even standard training paradigms (baseline mAP=46.6\%); 3) Bounding box localization demands intensive contextual semantics processing. The high coupling (0.78) observed likely reveals that the inherent features of object detection have entanglement across images.


\subsection{Threats to Validity}
\subsubsection{External validity}
Although we argue that NeMo is generalizable to various DNN models, it is unrealistic to evaluate it on all DNN models due to the huge time and computational costs. To mitigate the potential threats to generalizability, our extensive experiments use two mainstreams of DNN models including two Transformer-based and four CNN-based models.
Moreover, as in existing works~\cite{CNNPan,DNNPan,MwT,SeaM}, we evaluate NeMo on the computer vision task. The validity of NeMo on language models, particularly Transformer-based models like CodeBERT, remains to be investigated. 
Given that NeMo supports the modularization of Transformer-based vision models, it is potentially applicable to language models as well. We leave these investigations for future work.

\subsubsection{Internal validity}
One threat to internal validity may come from subject selection bias. To reduce this threat, we use CIFAR-10 and SVHN datasets, VGG16 and ResNet18 models from PyTorch~\cite{pytorch}, as well as ViT and DeiT models from HuggingFace~\cite{huggingface}, which are well organized and widely used.
Additionally, to mitigate threats from the stochastic nature of deep learning models, we repeated the ViT\_s model training experiments on SVHN 10 times using different random seeds. The resulting standard deviations across these experiments for Accuracy, NRR, Cohesion, and Coupling between experiments were 0.005, 0.0112, 0.0047, and 0.0181, respectively, demonstrating that \projectName's performance is stable and consistent.

\subsubsection{Construct validity}
The construct validity refers to the evaluation metrics we used. The metrics of cohesion and coupling of DNN modularization are proposed in MwT~\cite{MwT} and have been proven effective in evaluating the performance of DNN modularization approaches in modularity.

\subsection{Limitations and Future Work}
\subsubsection{Generative Model.}
Although \projectName's architecture exhibits compatibility with various models, its application to many other tasks remains highly challenging. For generative models such as GPTs, where sequences are produced through autoregressive output mechanisms, the neuron identifier is required to determine activated neurons for each token throughout the entire sequence. This process consequently incurs computational overhead several times greater than that of classification models. For other types of models such as detection, segmentation and reinforcement learning models, there are still limitations for \projectName. The inherent complexity arises from the fact that most tasks lack well-defined categorical boundaries, with neural units exhibiting tight functional interdependencies - attempts at modular decomposition often incur prohibitive performance degradation. Even for tasks like object detection, where bounding box annotation ostensibly operates within defined classes, the critical dependence on global contextual understanding fundamentally challenges \projectName. Building upon these challenges, our ongoing research prioritizes the development of label-agnostic modularization frameworks, with particular emphasis on their applicability to generative architectures and other mainstream models.

\subsubsection{Modular Training Overhead}
The integration of Neuron Identifier introduces measurable computational overhead in modular training, typically ranging from 30\% to 70\%, depending on model architectures and hardware configurations. Empirical evaluation reveals concrete manifestations: For ViT-based 10-class classification on CIFAR-10, standard training completes in approximately 5 hours versus modular training's 7 hours (40\% extra training time). This extra training time aligns with observations in MwT implementations, demonstrating consistent overhead patterns across different frameworks. While direct cross-architectural comparison between \projectName and MwT proves challenging due to their different model architectures, their relative overhead ratios are almost the same. Notably, when comparing the loss functions between \projectName and MwT on the same model, \projectName demonstrates accelerated convergence with fewer training epochs, highlighting its inherent optimization advantages.

Considering the additional training overhead, \projectName is suitable for the scenario, as discussed in Section \ref{sec:case_study}, where trained models will be extensively reused and specific functionalities are required for downstream tasks.
In this case, \projectName can reduce computational costs by decreasing the number of FLOPs through selective neuron reuse, eliminating redundant neurons.

\subsubsection{Large-scale Datasets}
The empirical observations reveal some challenges when applying \projectName to ViT architectures with large-scale datasets. While researchers try to improve the accuracy of ViT by using larger datasets, our attempts to scale training to ImagenNet datasets (1000 classes) exposed two fundamental constraints: (1) Training ViT on a large dataset with \projectName always needs days or weeks, and (2) Performance degradation emerges as the number of classes increases, and that is because, in each class, there are fewer neurons to allocate. Current mitigation strategies involving coupling weight relaxation and adaptive regularization only achieve partial remediation. 
Our future work will focus on the above issues in various models, including discriminative and generative models.

\section{Related Work}

The functionality of traditional software systems is modularized, making them easier to maintain and further develop. Similarly, from the perspective of software systems, a DNN model can be viewed as a ``complex system'' constructed in a data-driven manner and equipped with various functionalities. 
DNN modularization aims to decompose such a ``complex system'' with its functionalities into a set of sub-functionalities, thus facilitating the maintenance and development of DNN models.
To this end, the software engineering community has been exploring two primary directions: \textit{modularizing-after-training} and \textit{modularizing-while-training} techniques.

\subsection{Modularizing-after-training}
~\textit{Modularizing-after-training} methods focused on decomposing well-trained DNN models by identifying the groups of weights that are responsible for the corresponding classes~\cite{CNNSplitter, GradSplitter, CNNPan, DNNPan, NUSxinchao2022deep, NUSxinchao2023partial, SeaM}.
Specifically, Pan et al. pioneered DNN modularization~\cite{DNNPan}, decomposing multi-class classification fully connected neural network (FCNN) models into modules, with each module identifying a single class.
This approach determines the relevance of the weights to a specific class based on neuron activation and sets irrelevant weights to zero, resulting in modules with a single functionality but the same size as the model.
Their subsequent work extended the neuron activation-based idea to the modularization of CNNs~\cite{CNNPan} and LSTMs ~\cite{iowaRNN}.
Since these approaches measure the relevance at the individual weight level and do not remove irrelevant weights physically, we classify them as unstructured modularization.
In contrast,
Qi et al. were the first to
propose structured modularization approaches for CNN models, including CNNSplitter~\cite{CNNSplitter} and GradSplitter~\cite{GradSplitter}. These approach decompose trained CNN models by searching relevant convolutional kernels (a special structural component in CNNs) and physically removing the irrelevant ones, thereby producing smaller modules.
However, since the model is not explicitly trained for modularization, interdependencies between model weights and high coupling between weights significantly limit the efficiency and effectiveness of modularizing-after-training techniques~\cite{MwT}. 

\subsection{Modularizing-while-training}
To address the above limitations, Qi et al. proposed a new paradigm for DNN modularization, \textit{modularizing-while-training} (MwT)~\cite{MwT}. MwT aims to identify and optimize convolution kernels in CNNs for different functionalities during the training stage. 
It introduces the concepts of high cohesion and low coupling for DNN modularization and incorporates the designed cohesion and coupling losses to optimize these properties throughout the training process.  
By integrating modularization into the training phase, MwT achieves significant improvements in both efficiency (i.e., time cost of modularization) and effectiveness (i.e., module classification performance and size). 
However, MwT is implemented solely for CNN models and cannot be directly applied to Transformer-based architectures like ViT. 
Given the widespread success of Transformer-based models, 
we propose \projectName, which effectively addresses this challenge by extending robust modularization capabilities to Transformer-based models.

\section{Conclusion}
In this work, to overcome the difficulty of modular training and decomposition of the vision transformer model, we propose a novel neuron-level modularizing-while-training framework, \projectName, that achieves modular training and structured modularization for the Transformer-based models and can be easily extended to various DNN models.  
Additionally, we optimized the calculation of cohesion and coupling losses based on a contrastive learning approach, significantly reducing the convolution kernel and neuron retention rates while improving model accuracy. On-demand reuse experiments demonstrate that our method achieves better results in both Transformer-based and CNN models for on-demand reuse tasks. Furthermore, the improvements based on the contrastive learning method result in a loss function with only one adjustable hyperparameter, which is negatively correlated with both accuracy and NRR/KRR. This makes it easier for the algorithm to adapt to new models and achieve better performance.

In the future, we will extend \projectName to more model structures and achieve model decomposition structurally. Additionally, we will explore more on-demand reuse scenarios and build a more efficient approach to reduce the overhead of modular training.

The source code of \projectName and the experimental results are available at \url{https://github.com/XiaohanBi-Hub/NeMo}.

\section*{Acknowledgement}
This work was supported by National Key Research and Development Program of China under Grant No. 2024YFB3309602, National Natural Science Foundation
of China under Grant No. 62472017, Guangxi Collaborative Innovation Center of Multi-source Information Integration and Intelligent, and the Major Key Project of Peng Cheng Laboratory PCL2023A09.
\bibliographystyle{ACM-Reference-Format}
\bibliography{reference}

\end{document}